# A Novel Technique for Robust Training of Deep Networks With Multisource Weak Labeled Remote Sensing Data

Gianmarco Perantoni, and Lorenzo Bruzzone, *Fellow, IEEE*

*Abstract*—Deep learning has gained broad interest in remote sensing image scene classification thanks to the effectiveness of deep neural networks in extracting the semantics from complex data. However, deep networks require large amounts of training samples to obtain good generalization capabilities and are sensitive to errors in the training labels. This is a problem in remote sensing since highly reliable labels can be obtained at high costs and in limited amount. However, many sources of less reliable labeled data are available, e.g., obsolete digital maps. In order to train deep networks with larger datasets, we propose both the combination of single or multiple weak sources of labeled data with a small but reliable dataset to generate multisource labeled datasets and a novel training strategy where the reliability of each source is taken in consideration. This is done by exploiting the transition matrices describing the statistics of the errors of each source. The transition matrices are embedded into the labels and used during the training process to weigh each label according to the related source. The proposed method acts as a weighting scheme at gradient level, where each instance contributes with different weights to the optimization of different classes. The effectiveness of the proposed method is validated by experiments on different datasets. The results proved the robustness and capability of leveraging on unreliable source of labels of the proposed method.

## I. INTRODUCTION

IN this new era led by artificial intelligence (AI), deep learning (DL) architectures showed great capabilities in extracting the semantics from multisource data and deep convolutional neural networks (DCNNs) showed promising performance for remote sensing image scene classification [1]. However, DCNNs architectures are characterized by a large number of parameters to estimate. Hence, a large amount of training instances is required in order to achieve good generalization capabilities. On the one hand, the acquisition of a large amount of data is not a problem itself, e.g. the ESA's Copernicus programme has already produced huge amounts of data. On the other hand, the high cost for the acquisition of labels for satellite images is a well-known problem in the remote sensing community. Common solutions rely on specific learning strategies (e.g., data augmentation) and the use of pretrained models [2], which allow the use of smaller training sets. However, data augmentation is limited by the original labeled data and the pretrained models often are borrowed from other application domains, like computer vision, whose data are characterized by different physical properties, e.g., constraining the used spectral information to the RGB domain. Therefore, the desired optimal solution consists in the collection of a sufficiently large labeled dataset to train dedicated network models specific for the different kinds of remote sensing data.

Ideally, labels can be gathered by means of: 1) ground reference data collection; or 2) photointerpretation by experts. Both these methods allow to collect a limited amount of labeled data far from the required quantity for the training of deep architectures. However, other strategies do exist for the collection of labels. Remote sensing data are georeferenced, meaning that each pixel corresponds to a spatial location. This allows to exploit other auxiliary information sources derived from citizen sensor data or available digital maps [2]. However, these sources of labels are subject to errors, i.e., some of the labels are wrong, and thus can mislead the training of a machine learning model [3]. For this reason, they are usually avoided for training.

Weak labeled data sources could be exploited given the knowledge of the type of errors that can occur. One of the most common sources of error is the *label obsolescence*, which is related to the fact that the land cover can change over time and the given labels may refer to the past. It can be noted that, although to a lesser extent, this problem is also present with labeled data commonly considered reliable. However, in such cases this problem can be neglected. Other known sources of errors are inaccurate labelling, semantic inconsistency between

The authors are with the Department of Information Engineering and Computer Science, University of Trento, 38123 Trento, Italy (e-mail: gianmarco.perantoni@unitn.it; lorenzo.bruzzone@unitn.it).

different sources of labels, and geolocation errors [2].

In the literature, the presence of errors in the labels is usually modelled as noise. Noisy labels are known to be harmful for deep neural networks (DNNs) [4] and several techniques have been developed to solve this problem. These solutions can be divided into two classes: 1) techniques that try to first identify wrongly labeled data and then either to remove or to rectify them [5]–[9]; and 2) techniques that are designed to consider the presence of label noise during the training, like those based on either robust loss functions [4], [10]–[16], or the explicit modelling of label noise to reduce its negative effects during the training [17], [18].

Since inaccurate labels are relatively easy to collect in remote sensing, these techniques may be useful to train DNNs with larger datasets. However, on the one hand, the cleansing algorithms struggle to distinguish informative true labels from the noisy ones, and the resulting dataset may lose interesting samples or still be affected by label noise. On the other hand, techniques that use all the available data and keep in consideration the presence of noisy labels during the training can effectively exploit the larger quantity of training data.

To overcome the aforementioned problem of the reduced availability of reliable training labeled data, we explore the possibility of enlarging an otherwise small but reliable (clean) dataset with data labeled by single or multiple inaccurate/obsolete sources, each one labelling a different set of data. Indeed, from the perspective of satellite image classification, several unreliable (weak) sources of labels can be identified, e.g., by means of different obsolete digital maps. The exploitation of multiple sources of information is a key research topic in remote sensing [19]. Typically, this is done through data fusion, where different views are merged, thus allowing to extract more information than when using any individual data source. Similarly, here the idea is that more sources of labels can be used to produce a large dataset, thus generating a multisource labeled dataset. However, up to our knowledge none of the techniques developed so far to address the label noise problem has been considered under the multi-source labels setting. On top of that, each source is characterized by its own reliability and type of error in the labels, thus the reliability of each source should be taken in consideration during the training process [2]. In this context, we propose a novel training strategy capable of leveraging on the weak sources of labels and to weigh and exploit the given weak labels based on the knowledge of the related source. Specifically, similarly to [17], the errors in the labels are modelled for each source by a transition matrix that describes how true labels switch to weak labels. One should note that some types of error (e.g., land-cover changes) are related to variations in the underlying signal rather than to the presence of noise in the labels. Nonetheless, the use of transition matrices can address the problem by combining the noise model with the transition probabilities due to land-cover changes.

The novel contributions of this article can be summarized as follows:

1) To address the problem of the scarcity of labeled training data in remote sensing, a novel multisource approach is proposed, where one or multiple weak sources of labels are considered in addition to a small clean labeled dataset. By embedding information about the sources into the labels, the training procedure accordingly weighs each source and alleviates the negative effects of inaccurate labels. Furthermore, the proposed method is combined with robust loss functions to enhance the performance.
2) The proposed training approach is analyzed theoretically at gradient level, then evaluated and compared with standard training strategies in a simulated environment (i.e., artificial insertion of realistic errors) using two different benchmark datasets, where one and three weak sources are considered in addition to the clean one.
3) The proposed training strategy is general and can be used with any DL model. Thus, its applicability goes beyond the scenario studied here and the approach can be used for the training of DL models in any application context where data can be labeled by different unreliable sources.

The rest of this article is organized as follows. Related work is briefly reviewed in Section II. Section III introduces and analyzes the proposed multisource training strategy. Section IV describes the simulated environment and the design of experiments. Section V reports and discusses the experimental results. Finally, Section VI summarizes the article and suggests future works.

## II. RELATED WORKS

One of the main research efforts for training DL models using weak labeled data is about the design of specific loss functions that show desirable features (e.g., error tolerance). However, only recently this topic started to be studied in remote sensing [6], [14]–[16], [18]. Most of the studies come from the computer vision applications, which have a different context from remote sensing. Typically, in computer vision there is a single source of weak labeled data and the errors are caused only by some labelling noise, thus this problem is mostly referred to as *label noise*. Theoretical studies [4] showed that loss functions are robust against label noise when they satisfy a symmetry constraint or are bounded. These studies proved that the common *Categorical Cross Entropy* (CCE) is sensitive to label noise and instead the impractical *Mean Absolute Error* (MAE) is robust. However, the design of loss functions that satisfy the aforementioned constraints showed to be difficult. Thus, the research moved towards the design of losses with a behavior similar to MAE. One of these is the *Generalized Cross Entropy* (GCE) [11], which is a parametrized loss where CCE and MAE are extreme cases of the parameter's value. *Symmetric Learning* (SL) [10] combines the CCE with a symmetric counterpart similar to MAE in its implementation. Studies in [11] and [13] analyzed the robustness of loss functions from the gradient magnitude perspective, showing that robust loss functions can be understood as weighting schemes for the gradient of samples involved in the model parameters update. While CCE weights more the gradients of the disagreeing predictions, MAE considers also the confidence level of the classifier (i.e., the predicted probability), thus allowing the model to avoid an excessive overfitting of the given (weak) labels. Inspired by this, the *Improved MAE* (IMAE) was defined in [13]. It enhances the fitting capabilities of MAE while preserving its robustness, thus increasing its practical value.

In the context of remote sensing, few works trying to defining robust loss functions can be found in the literature. In [14] a robust loss function is defined based on an entropic optimal transport concept. In [15], a deep metric learning loss inspired on GCE and the normalized softmax loss is proposed, showing good performances on different tasks including classification, clustering, and retrieval. The same authors also proposed another deep metric learning loss [16] based on the maximization of the leave-one-out K-NN score for uncovering the inherent neighborhood structure among the images in feature space, where the loss function down-weights the potentially wrongly labeled images by pruning those with a low leave-one-out K-NN score.

Another approach to the training of DL models using weak labeled data is the explicit modelling of the label noise, which is then exploited to mitigate its effects on the training process. Typically, this is done by assuming class-dependent noise, which allows to characterize the error using the transition probabilities between the true and the noisy labels. Thus, a noise transition matrix can be defined. However, this matrix has to be estimated, usually requiring a set of reliable labeled data. This can be achieved by a weak classifier trained with a small reliable training set and used to compare its predictions with the noisy labels in order to estimate the transition probabilities. In [17] two loss correction approaches are proposed: *forward* and *backward*. The *forward* approach exploits the transition probabilities to produce a noisy estimate of the output of the *softmax* classifier, which is then passed to the CCE loss. The *backward* approach computes first the loss values associated to all the available classes and then exploits the inverse transition matrix to produce a linear combination of all the possible losses, thus optimizing multiple classes at time. In [18], an approach similar to *forward* is adopted for road extraction from remote sensing images where the image variable is integrated into the noise model. However, the extension of this work to multiclass image scene classification may be troublesome.

In this work, we extend the *forward* loss correction approach to the multi-source labels setting, combine it with robust loss functions, and study its behavior at gradient level.

### III. PROPOSED APPROACH

#### A. Mathematical notation

Let us denote column vectors by lowercase bold letters (e.g., $\mathbf{v}$) and matrices by uppercase bold letters (e.g., $\mathbf{M}$). The *i*th component of a vector $\mathbf{v}$ is denoted as $v_i$, while the *i*th row and *j*th column of a matrix $\mathbf{M}$ are denoted as $\mathbf{M}_{i,\cdot}$ and $\mathbf{M}_{\cdot,j}$, respectively. The element of $\mathbf{M}$ in the *i*th row and *j*th column is denoted by $M_{i,j}$. Then, we define an input instance $X$ through its feature vector $\mathbf{x} \in \mathbb{R}^d$, and the associated label $k$ through the one-hot encoding vector $\mathbf{y} = \mathbf{e}^k$, $k \in \{1,2,\ldots,c\}$, where $c$ is the number of classes. Let $\mathcal{D} \doteq \{(\mathbf{x}^{(i)}, \mathbf{y}^{(i)}), i=1,2,\ldots,m\}$ be a dataset, where *m* is the number of instances in the dataset. Since the proposed training approach is general, we analyze it considering a generic DNN that can be substituted by any designed model. We only constrain the final layer to be a *softmax* function due its importance in our theoretical analysis. Therefore, a DNN for a *c*-class classification problem can be represented as a function $\mathbf{h}: \mathbb{R}^d \to \mathbb{R}^c$ with parameters $\theta$ that maps a feature vector $\mathbf{x}$ to a score vector $\mathbf{h}(\mathbf{x}, \theta) \in \mathbb{R}^c$ containing the scores for each class. Then, the posterior probabilities $p(\mathbf{y}|\mathbf{x})$ can be estimated by passing $\mathbf{h}(\mathbf{x},\theta)$ to the *softmax* function in order to obtain a vector $\hat{p}(\mathbf{y}|\mathbf{x},\theta) = \mathbf{u}(\mathbf{x},\theta) \in \Delta^{c-1}$, where $\Delta^{c-1} \subset [0,1]^c$ is the *c*-dimensional simplex. The components of $\mathbf{u}$ are denoted as follows:

$$u_i(\mathbf{x},\theta) = \frac{\exp[h_i(\mathbf{x},\theta)]}{\sum_{j=1}^c \exp[h_j(\mathbf{x},\theta)]} = \hat{p}(\mathbf{y}=\mathbf{e}^i|\mathbf{x},\theta). \quad (1)$$

#### B. Problem Formulation

Let us consider a loss functions $\ell[\mathbf{y},\mathbf{u}]$, where $\mathbf{y}$ is the one-hot target vector and $\mathbf{u}$ the predicted posterior probabilities of the classes, with the following property:

$$\ell[\mathbf{e}^k, \mathbf{u}] = f(u_k) \quad (2)$$

where $f:(0,1) \to \mathbb{R}$ is a continuous function. When combined with the loss correction, such loss functions show nice behaviors at gradient level, which will be analyzed later. Examples of loss functions that satisfy (2) are the CCE, MAE, SL [10] and GCE [11]:

$$CCE[\mathbf{e}^k, \mathbf{u}] = -\log(u_k)$$

$$MAE[\mathbf{e}^k, \mathbf{u}] = 2(1-u_k)$$

$$SL_{\alpha,\beta,A}[\mathbf{e}^k, \mathbf{u}] = \alpha CCE[\mathbf{e}^k, \mathbf{u}] + \beta \frac{A}{-2} MAE[\mathbf{e}^k, \mathbf{u}] \quad (3)$$

$$GCE_q[\mathbf{e}^k, \mathbf{u}] = \frac{1-u_k^q}{q}$$

where $\alpha,\beta > 0$, $A < 0$, and $q \in (0,1]$. For the gradient analysis we use the following convention:

$$\frac{d}{d\mathbf{x}} f = (\nabla_\mathbf{x} f)^\top \quad (4)$$

where $\nabla_\mathbf{x} f$ is the gradient of $f$ w.r.t. $\mathbf{x}$ denoted as a column vector.

Let us identify the sources of labeled data with positive numbers $s = 0,1,\ldots,S$, where $s = 0$ refers to the clean source and $S$ is the number of weak sources. Let $\tilde{\mathbf{y}}$ be the weak labels, whereas $\mathbf{y}$ refers to the clean ones. Each source $s$ is assumed to observe samples from the following distribution:

$$p(\mathbf{x},\tilde{\mathbf{y}}|s) = \sum_\mathbf{y} p(\tilde{\mathbf{y}}|\mathbf{y},s) p(\mathbf{y}|\mathbf{x}) p(\mathbf{x}) \quad (5)$$

where $p(\tilde{\mathbf{y}}|\mathbf{y},s)$ is the transition (conditional) probability given the true label $\mathbf{y}$ and the source *s*. Thus, let $\mathbf{T}^{(s)} \in [0,1]^{c \times c}$ be the transition matrix associated to source *s* specifying the probability of one label being flipped to another, so that
$$T_{j,k}^{(s)} \doteq p(\tilde{\mathbf{y}} = \mathbf{e}^k \mid \mathbf{y} = \mathbf{e}^j, s). \tag{6}$$

It is possible to define the overall (balanced) error rate $\eta^{(s)}$ of each source $s$ as follows:
$$\eta^{(s)} \doteq 1 - \frac{1}{c}\sum_{j=1}^{c} T_{j,j}^{(s)}. \tag{7}$$

Note that $\mathbf{T}^{(0)} = \mathbf{I}$, where $\mathbf{I}$ is the identity matrix, since $s = 0$ refers to the clean source. Let $\mathcal{D}_{train}^{(s)} = \{(\mathbf{x}^{(s,i)}, \tilde{\mathbf{y}}^{(s,i)}), i = 1,2,\ldots,m^{(s)}\}$ be the set of (weak) labeled instances associated to the source $s$, where $m^{(s)}$ is the number of labeled instances of such source. Hence, the total number of labeled instances is $m = \sum_{s=0}^{S} m^{(s)}$. Then, let $\mathcal{D}_{train}^{MS} = \bigcup_{s=0}^{S} \mathcal{D}_{train}^{(s)}$ be the multisource dataset obtained by merging the labeled data of all the sources. A standard training strategy would be the minimization of a modified expected loss, where each sample is weighted according to the related label source:
$$\mathcal{L}_{\mathcal{D}_{train}^{MS}}^{std}(\theta) = \frac{1}{m}\sum_{s=0}^{S} \omega^{(s)} \sum_{i=1}^{m^{(s)}} \ell[\mathbf{y}^{(s,i)}, \mathbf{u}(\mathbf{x}^{(s,i)},\theta)] \tag{8}$$

where $\omega^{(s)}$ is the weight associated to source $s$, chosen to reflect the reliability of the source. However, such strategy has several problems. Even though the weak labeled samples are weighted, the training instance is still used to optimize the class associated to the weak label without considering the likelihood associated to other classes. Then, any robust loss function in the form of (2) acts by weighting the gradient associated to the weak label, but still it does not consider the other classes. This behavior will be proven later. On the contrary, the proposed approach acts as a complex weighting scheme where the same sample is used to optimize multiple classes, weighted on the basis of the type of error of the current labelling source and the confidence level of the prediction.

### C. Transition matrix estimation

The proposed training method relies on the estimation of the transition matrices of each source in order to characterize them. Ideally, a pool of clean-to-weak label pairs is necessary in order to estimate every transition probability. However, on the one hand, this requires to collect both the true and the weak labels of a set of instances for each source and it is usually a troublesome task. On the other hand, sacrificing on the estimation quality, quasi-clean labels can be easier to collect, e.g., as the predictions of a classifier trained on a small clean labeled dataset. In this work we assume that some clean labeled data are available. Indeed, as previously stated, ground reference data collection or photointerpretation by experts are strategies that allow the collection of a limited but reliable set of labels. Thus, the transition matrices can be estimated by: 1) training a baseline classifier with the small clean dataset $\mathcal{D}_{train}^{(0)}$; 2) using the trained classifier to compute the confusion matrices $\mathbf{C}^{(s)}$ between the baseline and each dataset $\mathcal{D}_{train}^{(s)}$, $s = 1,\ldots,S$; and 3) using the confusion matrices to compute approximate estimations $\hat{\mathbf{T}}^{(s)}$ of the transition matrices $\mathbf{T}^{(s)}$. Note that even if the baseline classifier is trained with few instances, in this step we do not need high quality estimations as long as the underlying relationships between classes are captured.

### D. Proposed multisource training approach

Instead of a standard training approach as in (8), we chose to explore a loss correction approach [17] and thus to keep $\omega^{(s)} = 1 \; \forall s$. The idea is that, on the one hand, training a model with weak labels and a standard training strategy would estimate a predictor for weak labels $\mathbf{u}(\mathbf{x},\theta) = \hat{p}(\tilde{\mathbf{y}} \mid \mathbf{x}, \theta)$. On the other hand, making explicit the dependencies between weak and true labels as in (5) would allow to train a predictor for true labels. Thus, the proposed approach modifies the predicted probabilities $\mathbf{u}$ for each sample $i$ of each weak source $s$ $\mathbf{x}^{(s,i)}$, $s \in \{0,1,\ldots,S\}$, $i \in \{1,2,\ldots,m^{(s)}\}$, according to the transition matrix $\hat{\mathbf{T}}^{(s)}$ as follows:

$$\begin{aligned}\tilde{u}_k(\mathbf{x}^{(s,i)},\theta) &= \hat{p}(\tilde{\mathbf{y}} = \mathbf{e}^k \mid \mathbf{x}^{(s,i)},\theta) \\ &= \sum_{j=1}^{c} \hat{p}(\tilde{\mathbf{y}} = \mathbf{e}^k \mid \mathbf{y} = \mathbf{e}^j, s)\hat{p}(\mathbf{y} = \mathbf{e}^j \mid \mathbf{x}^{(s,i)},\theta) \\ &= \sum_{j=1}^{c} \hat{T}_{j,k}^{(s)} u_j(\mathbf{x}^{(s,i)},\theta). \end{aligned} \tag{9}$$

In vector form this can be written as follows:
$$\tilde{\mathbf{u}}(\mathbf{x}^{(s,i)},\theta) = \hat{\mathbf{T}}^{(s)\top}\mathbf{u}(\mathbf{x}^{(s,i)},\theta). \tag{10}$$

Therefore, the modified expected loss becomes the following:
$$\begin{aligned}\mathcal{L}_{\mathcal{D}_{train}^{MS}}(\theta) &= \frac{1}{m}\sum_{s=0}^{S}\sum_{i=1}^{m^{(s)}} \ell[\mathbf{y}^{(s,i)}, \tilde{\mathbf{u}}(\mathbf{x}^{(s,i)},\theta)] \\ &= \frac{1}{m}\sum_{s=0}^{S}\sum_{i=1}^{m^{(s)}} \ell[\mathbf{y}^{(s,i)}, \hat{\mathbf{T}}^{(s)\top}\mathbf{u}(\mathbf{x}^{(s,i)},\theta)].\end{aligned} \tag{11}$$

### E. Gradient analysis

While in (11) there is no explicit weighting of the training instances, an implicit weighting is present at the level of the gradient (see Fig. 1). Let us consider a generic instance $\mathbf{x}$ with weak label $\mathbf{y} = \mathbf{e}^k$ generated by a weak source with estimated transition matrix $\hat{\mathbf{T}}$. The gradient of the loss function $\ell$ w.r.t. the model parameters $\theta$ can be split into two terms:
$$\frac{d}{d\theta}\ell[\mathbf{e}^k, \tilde{\mathbf{u}}(\mathbf{x},\theta)] = \frac{d}{d\mathbf{h}}\ell[\mathbf{e}^k, \tilde{\mathbf{u}}]\frac{d}{d\theta}\mathbf{h}(\mathbf{x},\theta). \tag{12}$$

The second term can be seen as the actual gradient of the model associated to the instance $\mathbf{x}$, while the first one can be considered as a complex weight $\boldsymbol{\omega}(\hat{\mathbf{T}}, \mathbf{e}^k, \mathbf{u})$ for that gradient, whose value is calculated based on the source, the given weak label, and the model prediction:
$$\begin{aligned}\boldsymbol{\omega}(\hat{\mathbf{T}}, \mathbf{e}^k, \mathbf{u}) &= \nabla_{\mathbf{h}}\ell[\mathbf{e}^k, \tilde{\mathbf{u}}] = \left(\frac{d}{d\mathbf{h}}\ell[\mathbf{e}^k, \tilde{\mathbf{u}}]\right)^\top \\ \frac{d}{d\theta}\ell[\mathbf{e}^k, \tilde{\mathbf{u}}(\mathbf{x},\theta)] &= \boldsymbol{\omega}(\hat{\mathbf{T}}, \mathbf{e}^k, \mathbf{u})^\top \frac{d}{d\theta}\mathbf{h}(\mathbf{x},\theta).\end{aligned} \tag{13}$$

Recalling (10), the weighting term $\boldsymbol{\omega}(\hat{\mathbf{T}}, \mathbf{e}^k, \mathbf{u})$ of the gradient can be expanded as follows:

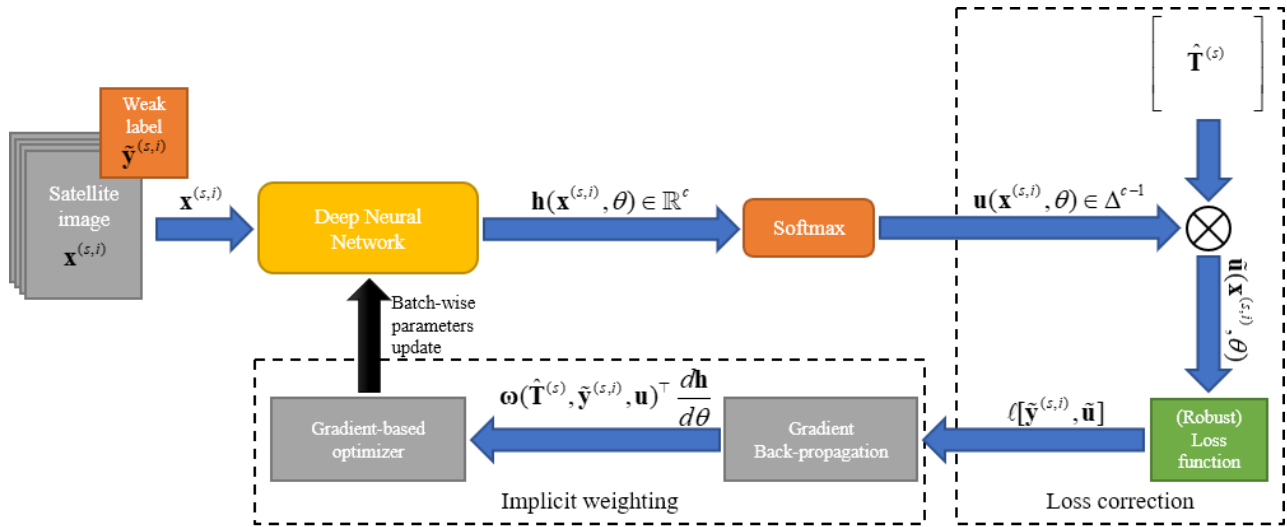

Fig. 1. Flowchart of the proposed training strategy as applied to satellite image scene classification. The Loss correction step induces an implicit weighting of the gradient associated to the current sample, based on the transition matrix of the source involved, the given weak label and the current predictions of the DNN.

$$\boldsymbol{\omega}(\hat{\mathbf{T}}, \mathbf{e}^k, \mathbf{u})^\top = \frac{d}{d\tilde{\mathbf{u}}} \ell[\mathbf{e}^k, \tilde{\mathbf{u}}] \frac{d}{d\mathbf{u}} \tilde{\mathbf{u}} \frac{d}{d\mathbf{h}} \mathbf{u}$$

$$= \frac{d}{d\tilde{\mathbf{u}}} \ell[\mathbf{e}^k, \tilde{\mathbf{u}}] \hat{\mathbf{T}}^\top \begin{bmatrix} \frac{d}{d\mathbf{h}} u_1 \\ \cdots \\ \frac{d}{d\mathbf{h}} u_c \end{bmatrix} \quad (14)$$

where $\frac{d}{d\mathbf{u}}\tilde{\mathbf{u}}$ and $\frac{d}{d\mathbf{h}}\mathbf{u}$ are Jacobians. Considering a loss function with the property of (2), by using (4) the weighting term can be simplified as follows:

$$\boldsymbol{\omega}(\hat{\mathbf{T}}, \mathbf{e}^k, \mathbf{u}) = \frac{\partial}{\partial \tilde{u}_k} \ell[\mathbf{e}^k, \tilde{\mathbf{u}}] \sum_{j=1}^{c} \hat{T}_{j,k} \nabla_\mathbf{h} u_j . \quad (15)$$

It is straightforward to see that for $\hat{\mathbf{T}} = \mathbf{I}$ the gradient reduces to the standard case, where utmost faith is given to the training labels. For comparison, the standard weighting term of (8) is:

$$\boldsymbol{\omega}^{std}(s, \mathbf{e}^k, \mathbf{u}) = \frac{\partial}{\partial u_k} \ell[\mathbf{e}^k, \mathbf{u}] \omega^{(s)} \nabla_\mathbf{h} u_k . \quad (16)$$

One can see that (15) and (16) share a scalar term, i.e., the derivative of the loss, which acts as a scalar weight that adjust the magnitude of the gradient. For example, CCE would give larger weights to gradients of disagreeing predictions, whereas MAE would show no preferences since its derivative is a constant [11]. Then, to better understand the differences between the two strategies, one should focus on the role of the gradient of the *softmax* functions:

$$\nabla_\mathbf{h} u_j = u_j \left( \mathbf{e}^j - \mathbf{u} \right)$$
$$= \mathbf{e}^j \odot \mathbf{u} - u_j \mathbf{u} \quad (17)$$

where $\odot$ is the Hadamard product. One can see that, for a standard training strategy, if the given label is $k$, then the gradient involved is $\nabla_\mathbf{h} u_k$, which allows to optimize class $k$ since it would make the score $h_k$ increase while decreasing the other scores $h_j, j \neq k$. Hence, any loss function as (2) embedded in a standard training strategy would only consider the optimization of the class of the given label, regardless of the likelihood of any other class, as one can see in (16). Instead, from (15) we can observe that the proposed approach optimizes multiple classes at time by considering the likelihood, i.e., $\hat{T}_{j,k}$, of any class $j$ being the true class flipped to class $k$. Note that in (15) the gradients of the *softmax* receive the weights from the columns of $\hat{\mathbf{T}}$, which is row-stochastic. Hence, the weights $\hat{T}_{j,k}$ are not normalized.

Now, let us consider the GCE loss (3) due to its property of generalization of both CCE and MAE (up to a constant of proportionality) for $q \to 0$ and $q = 1$, respectively:

$$\frac{\partial}{\partial u_k} GCE_q[\mathbf{e}^k, \mathbf{u}] = -u_k^{q-1} . \quad (18)$$

It can be shown that by combining the gradients of the *softmax* functions (17) in the weighting term (15), and considering the GCE loss (18), one obtains the following:

$$\boldsymbol{\omega}_{GCE}(\hat{\mathbf{T}}, \mathbf{e}^k, \mathbf{u}) = -\tilde{u}_k^{q-1} \left( \hat{\mathbf{T}}_{\cdot,k} \odot \mathbf{u} - \tilde{u}_k \mathbf{u} \right)$$
$$= -\tilde{u}_k^{q} \left( \frac{\hat{\mathbf{T}}_{\cdot,k} \odot \mathbf{u}}{\tilde{u}_k} - \mathbf{u} \right). \quad (19)$$

To better understand which classes are being optimized, let us analyze which components of (19) are negative. It can be shown that class $j$ is optimized when $u_j \neq 0$ and $\hat{T}_{j,k} > \tilde{u}_k = \sum_i \hat{T}_{i,k} u_i$. Therefore, since $\tilde{u}_k$ is a weighted average of the components of $\hat{\mathbf{T}}_{\cdot,k}$, the most likely prior true label $j_{max} = \arg\max_j(\hat{T}_{j,k})$ is always optimized, given that $u_{j_{max}} \neq 0$, whereas the least likely class $j_{min} = \arg\min_j(\hat{T}_{j,k})$ is never optimized. Thus, it is sufficient for the estimated transition matrices to have the maximum of each column in the same positions as in the true transition matrices, i.e., $\arg\max_j(\hat{T}_{j,k}) = \arg\max_j(T_{j,k}) \; \forall k$, in order to allow the model to learn under the majority voting

(a) 

$$\begin{array}{c}\text{Ann. Crop}\\\text{Forest}\\\text{Herb. Veg.}\\\text{Highway}\\\text{Industrial}\\\text{Pasture}\\\text{Perm. Crop}\\\text{Residential}\\\text{River}\\\text{Sea\&Lake}\end{array}\begin{bmatrix}1-\frac{\eta}{0.8} & 0 & 0 & 0 & 0 & \frac{\eta}{2\cdot0.8} & \frac{\eta}{2\cdot0.8} & 0 & 0 & 0\\ 0 & 1-\frac{\eta}{0.8} & \frac{\eta}{0.8} & 0 & 0 & 0 & 0 & 0 & 0 & 0\\ 0 & 0 & 1-\frac{\eta}{0.8} & 0 & 0 & \frac{\eta}{0.8} & 0 & 0 & 0 & 0\\ \frac{\eta}{2\cdot0.8} & \frac{\eta}{2\cdot0.8} & 0 & 1-\frac{\eta}{0.8} & 0 & 0 & 0 & 0 & 0 & 0\\ 0 & 0 & 0 & 0 & 1-\frac{\eta}{0.8} & \frac{\eta}{2\cdot0.8} & 0 & \frac{\eta}{2\cdot0.8} & 0 & 0\\ 0 & 0 & \frac{\eta}{0.8} & 0 & 0 & 1-\frac{\eta}{0.8} & 0 & 0 & 0 & 0\\ \frac{\eta}{2\cdot0.8} & 0 & 0 & 0 & 0 & \frac{\eta}{2\cdot0.8} & 1-\frac{\eta}{0.8} & 0 & 0 & 0\\ 0 & \frac{\eta}{2\cdot0.8} & 0 & 0 & \frac{\eta}{2\cdot0.8} & 0 & 0 & 1-\frac{\eta}{0.8} & 0 & 0\\ 0 & 0 & 0 & 0 & 0 & 0 & 0 & 0 & 1 & 0\\ 0 & 0 & 0 & 0 & 0 & 0 & 0 & 0 & 0 & 1\end{bmatrix}$$

(b)

$$\begin{bmatrix}1-\eta & \frac{\eta}{9} & \frac{\eta}{9} & \frac{\eta}{9} & \frac{\eta}{9} & \frac{\eta}{9} & \frac{\eta}{9} & \frac{\eta}{9} & \frac{\eta}{9} & \frac{\eta}{9}\\ \frac{\eta}{9} & 1-\eta & \frac{\eta}{9} & \frac{\eta}{9} & \frac{\eta}{9} & \frac{\eta}{9} & \frac{\eta}{9} & \frac{\eta}{9} & \frac{\eta}{9} & \frac{\eta}{9}\\ \frac{\eta}{9} & \frac{\eta}{9} & 1-\eta & \frac{\eta}{9} & \frac{\eta}{9} & \frac{\eta}{9} & \frac{\eta}{9} & \frac{\eta}{9} & \frac{\eta}{9} & \frac{\eta}{9}\\ \frac{\eta}{9} & \frac{\eta}{9} & \frac{\eta}{9} & 1-\eta & \frac{\eta}{9} & \frac{\eta}{9} & \frac{\eta}{9} & \frac{\eta}{9} & \frac{\eta}{9} & \frac{\eta}{9}\\ \frac{\eta}{9} & \frac{\eta}{9} & \frac{\eta}{9} & \frac{\eta}{9} & 1-\eta & \frac{\eta}{9} & \frac{\eta}{9} & \frac{\eta}{9} & \frac{\eta}{9} & \frac{\eta}{9}\\ \frac{\eta}{9} & \frac{\eta}{9} & \frac{\eta}{9} & \frac{\eta}{9} & \frac{\eta}{9} & 1-\eta & \frac{\eta}{9} & \frac{\eta}{9} & \frac{\eta}{9} & \frac{\eta}{9}\\ \frac{\eta}{9} & \frac{\eta}{9} & \frac{\eta}{9} & \frac{\eta}{9} & \frac{\eta}{9} & \frac{\eta}{9} & 1-\eta & \frac{\eta}{9} & \frac{\eta}{9} & \frac{\eta}{9}\\ \frac{\eta}{9} & \frac{\eta}{9} & \frac{\eta}{9} & \frac{\eta}{9} & \frac{\eta}{9} & \frac{\eta}{9} & \frac{\eta}{9} & 1-\eta & \frac{\eta}{9} & \frac{\eta}{9}\\ \frac{\eta}{9} & \frac{\eta}{9} & \frac{\eta}{9} & \frac{\eta}{9} & \frac{\eta}{9} & \frac{\eta}{9} & \frac{\eta}{9} & \frac{\eta}{9} & 1-\eta & \frac{\eta}{9}\\ \frac{\eta}{9} & \frac{\eta}{9} & \frac{\eta}{9} & \frac{\eta}{9} & \frac{\eta}{9} & \frac{\eta}{9} & \frac{\eta}{9} & \frac{\eta}{9} & \frac{\eta}{9} & 1-\eta\end{bmatrix}$$

(c)

$$\begin{array}{c}\text{Ann. Crop}\\\text{Forest}\\\text{Herb. Veg.}\\\text{Highway}\\\text{Industrial}\\\text{Pasture}\\\text{Perm. Crop}\\\text{Residential}\\\text{River}\\\text{Sea\&Lake}\end{array}\begin{bmatrix}1-\frac{\eta}{0.6} & \frac{\eta}{2\cdot0.6} & 0 & 0 & 0 & \frac{\eta}{2\cdot0.6} & 0 & 0 & 0 & 0\\ 0 & 1 & 0 & 0 & 0 & 0 & 0 & 0 & 0 & 0\\ 0 & 0 & 1 & 0 & 0 & 0 & 0 & 0 & 0 & 0\\ 0 & \frac{\eta}{2\cdot0.8} & \frac{\eta}{2\cdot0.6} & 1-\frac{\eta}{0.6} & 0 & 0 & 0 & 0 & 0 & 0\\ \frac{\eta}{3\cdot0.6} & \frac{\eta}{3\cdot0.6} & 0 & 0 & 1-\frac{\eta}{0.6} & \frac{\eta}{3\cdot0.6} & 0 & 0 & 0 & 0\\ 0 & 0 & \frac{\eta}{0.6} & 0 & 0 & 1-\frac{\eta}{0.6} & 0 & 0 & 0 & 0\\ 0 & \frac{\eta}{2\cdot0.6} & \frac{\eta}{2\cdot0.6} & 0 & 0 & 0 & 1-\frac{\eta}{0.6} & 0 & 0 & 0\\ 0 & \frac{\eta}{3\cdot0.6} & \frac{\eta}{3\cdot0.6} & 0 & 0 & \frac{\eta}{3\cdot0.6} & 0 & 1-\frac{\eta}{0.6} & 0 & 0\\ 0 & 0 & 0 & 0 & 0 & 0 & 0 & 0 & 1 & 0\\ 0 & 0 & 0 & 0 & 0 & 0 & 0 & 0 & 0 & 1\end{bmatrix}$$

(d)

$$\begin{bmatrix}1-\eta & 0 & 0 & \frac{\eta}{3} & 0 & \frac{\eta}{3} & \frac{\eta}{3} & 0 & 0 & 0\\ 0 & 1-\eta & \frac{\eta}{2} & \frac{\eta}{2} & 0 & 0 & 0 & 0 & 0 & 0\\ 0 & \eta & 1-\eta & 0 & 0 & 0 & 0 & 0 & 0 & 0\\ \frac{\eta}{3} & \frac{\eta}{3} & 0 & 1-\eta & 0 & 0 & \frac{\eta}{3} & 0 & 0 & 0\\ 0 & 0 & 0 & 0 & 1-\eta & 0 & 0 & \eta & 0 & 0\\ \frac{\eta}{2} & 0 & 0 & 0 & 0 & 1-\eta & \frac{\eta}{2} & 0 & 0 & 0\\ \frac{\eta}{3} & 0 & 0 & \frac{\eta}{3} & 0 & \frac{\eta}{3} & 1-\eta & 0 & 0 & 0\\ 0 & 0 & 0 & 0 & \eta & 0 & 0 & 1-\eta & 0 & 0\\ 0 & 0 & 0 & 0 & 0 & 0 & 0 & 0 & 1-\eta & \eta\\ 0 & 0 & 0 & 0 & 0 & 0 & 0 & 0 & \eta & 1-\eta\end{bmatrix}$$

Fig. 2. Transition matrix templates used with the EuroSAT dataset for the generation of the weak sources considered in the experiments, parametrized by the balanced error rate $\eta$. (a) Class-dependent errors due to both interclass similarities and land-cover changes. (b) Uniform errors: a less likely scenario in real world settings. (c) Class-dependent errors due to land-cover-changes. (d) Class-dependent errors due to interclass similarities.

assumption, regardless of the error rate $\eta$ (e.g., if labels are simply permuted, then $\hat{\mathbf{T}}$ would be close to the permutation matrix $\mathbf{T}$, allowing almost perfect recovery of weak labels even if $\eta$ is high). Note that a standard training approach with any robust loss functions with the property (2) allows to exploit the majority voting assumption only when $\mathbf{T}$ satisfies the following constraint [4]:

$$T_{j,j} > T_{j,k} \quad \forall j \; \forall k \neq j . \quad (20)$$

Otherwise, the optimization would be severely misled by the erroneous majority, i.e., the data mislabeled to class $\arg\max_k(\hat{T}_{j,k})$. Returning to the proposed training strategy, other classes may be optimized simultaneously to $j_{\max}$. Indeed, any class $j$ whose $\hat{T}_{j,k}$ is greater than $\tilde{u}_k$ is optimized. These properties allow to exploit the available data and suggest that the error rate is not as important as the entropy of $\mathbf{T}$ in affecting the performances of the proposed training strategy. Indeed, if the errors are uniformly spread among the classes, than the influence of the error rate on the performances increases, since there is no information in the weak labels that can be exploited. Nonetheless, this type of error is less common in real world settings when considering weak labeled data sources. Common sources of errors, e.g., land-cover changes and inaccurate labelling, cause weak labels that are dependent of the original label. For example, a water body has few chances of changing over time, while a wooded area may have been deforested and substituted by crops or a residential area. Therefore, for some classes an obsolete map may refer to the past and thus provide the wrong label. Another example is the use of inaccurate maps, which may suffer of the interclass similarities typical of satellite images, e.g., between residential and industrial areas, or between different crop types. Hence, the probabilities in the transition matrix would be collected on transitions between similar classes and not spread among all the classes.

From (19) one can note that $\left\| (\hat{\mathbf{T}}_{\bullet,k} \odot \mathbf{u})/\tilde{u}_k \right\|_1 = 1$, and since all the components are nonnegative, $(\hat{\mathbf{T}}_{\bullet,k} \odot \mathbf{u})/\tilde{u}_k$ can be viewed as a probability distribution. Hence, the L1-norm of the difference is bounded:

$$0 \leq \left\| \frac{\hat{\mathbf{T}}_{\bullet,k} \odot \mathbf{u}}{\tilde{u}_k} - \mathbf{u} \right\|_1 \leq 2 . \quad (21)$$

Let us consider a confident prediction $\mathbf{u} \approx \mathbf{e}^j$ and $\hat{T}_{j,k} \neq 0$. Then, $(\hat{\mathbf{T}}_{\bullet,k} \odot \mathbf{u})/\tilde{u}_k \approx \mathbf{u}$, and thus all the components of (19) tends to zero. Note that CCE (i.e., $q \to 0$ thus $\tilde{u}_k^q \to 1$) known to be sensitive to weak labels, now becomes robust against them since it produces small gradients for strongly confident and disagreeing predictions, given that $\hat{T}_{j,k} \neq 0$, meaning that these likely wrongly labeled samples will have a lower chance of misleading the optimization process. On the contrary, if $\hat{T}_{j,k} = 0$ then $\left\| (\hat{\mathbf{T}}_{\bullet,k} \odot \mathbf{u})/\tilde{u}_k - \mathbf{u} \right\|_1 \approx 2$. Thus, if there are no chances of $j$ to be the true label, then $\mathbf{u} \approx \mathbf{e}^j$ is treated as wrong, and the parameters are updated to decrease the score $h_j$. Note that in the case of MAE (i.e., $q = 1$ thus $\tilde{u}_k^q = \tilde{u}_k$) such a confident prediction would still lead to a small gradient magnitude.

## IV. DESCRIPTION OF DATASETS AND DESIGN OF EXPERIMENTS

This section presents the experimental setup. First, it illustrates the datasets used along with the related data pre-processing phase and DL models adopted. Then, the multisource dataset generation process and the design of the experiments are described.

### A. EuroSAT Dataset

In order to quantitatively assess the performances in a controlled manner, we mainly evaluated the proposed training strategy on a land-use/land-cover classification benchmark dataset, namely EuroSAT [20], [21]. The dataset is based on the openly and freely accessible Sentinel-2 multispectral satellite images (acquired in 13 different spectral bands) provided in the Copernicus Earth Observation program. The dataset consists of $c = 10$ classes and includes a total of $m = 27\,000$ labeled and geo-referenced images of shape 64×64 pixels. The spectral bands of the Sentinel-2 satellites have different spatial resolutions between 60m/pixel and 10m/pixel. In the dataset, bands with lower spatial resolution are upsampled to 10m/pixel. In the experiments, all the 13 spectral bands are used. The available classes are Annual Crop, Forest, Herbaceous Vegetation, Highway, Industrial buildings, Pasture, Permanent Crop, Residential buildings, River, and Sea & Lake. Each class contains 2000–3000 images.

The DL model chosen to be trained on the EuroSAT dataset is ResNet-50 [22], in its recent variant with pre-activation proposed in [23]. We modified the first convolutional layer in order to accept all the available spectral bands in input, thus increasing the total number of parameters of the DCNN. Consequently, the model is always trained from scratch. The great depth and the large number of parameters give to the DCNN the potential capacity to model the $p(\mathbf{y}|\mathbf{x})$ accurately. Note that the consequent higher chance of overfitting can be addressed by the larger quantity of labeled data made available by the weak sources. Table I shows the hyperparameters common to all the experiments with the EuroSAT dataset.

In the pre-processing phase, the mean value and the standard deviation are computed for each of the 13 spectral bands of the images in the training set. Then, they are used to normalize the images in both training and test sets.

Also, we considered the use of *data augmentation*. Data augmentation exploits the available labeled data to generate new synthetic data by applying specific transformations to the images, aiming to create a larger training dataset. Such transformations allow to insert prior knowledge into the dataset about the different ways and conditions the same target can be acquired, acting as a regularization technique being able to reinforce the generalization capabilities of the model. Since the proposed training strategy is not alternative to data augmentation, in our experiments we combined the two approaches. In particular, we considered simple transformations such as flips and rotations and explored their performances increase limit. Preliminary experiments showed that the adoption of all the possible combination of horizontal/vertical flips and 90-degree rotations was the limit beyond which the performances cease to achieve a considerable increase (e.g., applying rotations of any degree allows to generate a higher number of unique transformations, however the performances were similar to adopting all the possible 90-degree rotations). Hence, the augmented multisource labeled dataset results 8 times larger than the original one. Such offline data augmentation strategy has been used in all the experiments discussed in the following sections regarding the EuroSAT dataset. Note that in this way we can really assess the improvement in performance provided by the use of weak sources with the proposed training strategy using as reference the upper limit obtained by standard techniques with data augmentation.

TABLE I
HYPERPARAMETERS USED WITH RESNET-50 ON THE EUROSAT DATASET

| Parameter | Value |
|---|---|
| Initial learning rate | $1 \times 10^{-3}$ |
| Epochs | 60 |
| Batch size | 16 |
| Decay | $1 \times 10^{-6}$ |
| Momentum (Nesterov) | 0.9 |
| Optimizer | SGD |

### B. NWPU-RESISC45 Dataset

NWPU-RESISC45 [24] consists of $m = 31\,500$ remote sensing images covering $c = 45$ classes. Each class contains 700 RGB images with a shape of 256×256 pixels. The spatial resolution of this dataset varies from about 30m/pixel to 0.2m/pixel. This dataset was extracted from Google Earth by the experts in the field of remote sensing image interpretation.

Differently from EuroSAT, we chose this dataset both because it is widely used as benchmark and to prove the general applicability of the training strategy proposed here by adopting different data and different models. Being a dataset of RGB data, NWPU-RESISC45 can benefit from pre-trained models. This allows us to test the effectiveness of the proposed approach (which was devised for training from scratch a deep architecture) also in the case of pre-trained networks. We evaluated the proposed strategy on two models pre-trained on ImageNet [25]: ResNet-50 in its recent variant with pre-activation [23] and VGG16 [26]. Similar to [14], we replaced the last VGG16 layer with a two-layer MLP that maps to 512 hidden neurons before predicting the classes with $l_2 = 10^{-3}$ regularization, respectively. The ReLU activation function followed by a dropout layer with $p = 0.5$ are inserted before the last dense layer, which is then followed by a batch normalization layer before the softmax function. Similarly, we replaced the last ResNet-50 layer with a two-layer MLP that maps to 512 hidden neurons before predicting the classes (without L2 regularization). Only one batch normalization layer is inserted between the two dense layers before the ReLU activation. During the training, the two networks are fine-tuned

---

**Algorithm 1:** Multisource weak labeled data generation

**Inputs:**
SEED for the RNG
$\mathcal{D} = \{(\mathbf{x}^{(i)}, \mathbf{y}^{(i)}), i = 1, 2, \ldots, m\}$
$\mathbf{T}^{(s)}, s = 1, 2, \ldots, S$
$m^{(s)}, s = 0, 1, \ldots, S$

**Output:** Multisource labeled dataset $\mathcal{D}_{train}^{MS}$ and test set $\mathcal{D}_{test}$.

1: Shuffle $\mathcal{D}$.
2: Split $\mathcal{D}$ in $\mathcal{D}_{train}$ and $\mathcal{D}_{test}$, with $m_{train} = 0.8 \cdot m$ and $m_{test} = 0.2 \cdot m$, respectively.
3: Sample (without replacement) $m^{(0)}$ instances $(\mathbf{x}^{(0,i)}, \mathbf{y}^{(0,i)})$ from $\mathcal{D}_{train}$ for the clean source.
4: Create $\mathcal{D}_{train}^{(0)} = \{(\mathbf{x}^{(0,i)}, \mathbf{y}^{(0,i)}), i = 1, 2, \ldots, m^{(0)}\}$.
5: **for** $s = 1$ to $S$ **do**
6:   Sample (without replacement) $m^{(s)}$ instances $(\mathbf{x}^{(s,i)}, \mathbf{y}^{(s,i)})$ from $\mathcal{D}_{train}$ for the weak source $s$.
7:   **for** $i = 1$ to $m^{(s)}$ **do**
8:     Sample a new *weak* label $\tilde{\mathbf{y}}^{(s,i)}$ for $\mathbf{x}^{(s,i)}$ according to $p(\tilde{\mathbf{y}}^{(s,i)} | \mathbf{y}^{(s,i)}) = \mathbf{T}^{(s)\top} \mathbf{y}^{(s,i)}$.
9:   **end for**
10:   Create $\mathcal{D}_{train}^{(s)} = \{(\mathbf{x}^{(s,i)}, \tilde{\mathbf{y}}^{(s,i)}), i = 1, 2, \ldots, m^{(s)}\}$.
11: **end for**
12: Create $\mathcal{D}_{train}^{MS} = \bigcup_{s=0}^{S} \mathcal{D}_{train}^{(s)}$.
13: **return** $\mathcal{D}_{train}^{MS}$ and $\mathcal{D}_{test}$

---

by freezing the weights of the ResNet-50 and VGG16 layers. We optimized the different methods for 300 epochs using the SGD optimizer (learning rate of 0.01) with momentum equal to 0.9 using a batch size of 64 samples for VGG16 and of 128 samples for ResNet-50. Additionally, we used an early stopping criterion to terminate the training process, if the validation loss did not decrease for 25 epochs.

In the preprocessing phase we applied the ImageNet preprocessing steps used for pretraining the models: the input pixels are rescaled between -1 and 1 for ResNet, whereas they are converted from RGB to BGR and then channel-wise zero centered for VGG16.

### C. Multisource Weak Labeled Data Generation

Let $\mathcal{D} = \{(\mathbf{x}^{(i)}, \mathbf{y}^{(i)}), i = 1, 2, \ldots, m\}$ be a dataset, where we refer to images with a column vector $\mathbf{x}$ for simplicity and $\mathbf{y} = \mathbf{e}^j$, $j = 1, 2, \ldots, c$ is the one-hot vector encoding the label $j$ of sample $\mathbf{x}$. Since the benchmark datasets consist of highly reliable labeled data, the multisource weak labeled data are simulated by the artificial insertion of realistic errors into the given labels. In the experiments, subsets of $\mathcal{D}$ are sampled to generate the multisource training set $\mathcal{D}_{train}^{MS}$ and the test set $\mathcal{D}_{test}$. Algorithm 1 shows the process used to accomplish this task, once that the seed number for the random number generator (RNG, used for reproducible datasets) and the characteristics of each source (i.e., transition matrices and number of labeled instances) are defined. Note that (see line 8) for a generic sample $\mathbf{x}$ with true label $\mathbf{y} = \mathbf{e}^j$, the related source with transition matrix $\mathbf{T}$ would generate a weak label $\tilde{\mathbf{y}} = \mathbf{e}^k$ where $k$ is sampled according to the following probability distribution:

$$p(\tilde{\mathbf{y}} | \mathbf{y} = \mathbf{e}^j) = \mathbf{T}^\top \cdot \mathbf{y} = \mathbf{T}^\top \cdot \mathbf{e}^j = \mathbf{T}_{j,\cdot}. \quad (22)$$

Hence, the probability that the weak label $k$ is generated for the sample $\mathbf{x}$ when the true label is $j$ is $T_{j,k}$. Also note that $\mathbf{T}^{(0)} = \mathbf{I}$ since $\mathcal{D}_{train}^{(0)}$ is the known clean labeled dataset.

### D. Design of Experiments

To better assess the obtained results, most of the tests are run three times, each one on a different sampling of the data. In order to make these sampling reproducible, predefined seed numbers have been used for the RNG. Then, the mean Overall Accuracy (OA) on the test set $\mathcal{D}_{test}$ along with the standard deviation (unbiased estimator) have been reported. For each of the generated $\mathcal{D}_{train}^{MS}$, different combinations of training strategies and loss functions were considered:

1) Training strategy
   a) *Vanilla*: the standard training strategy for DNNs, where the DL model is trained by means of gradient-based batch-wise optimizers of the expected loss (8) without keeping in consideration the reliability of the sources (i.e., $\omega^{(s)} = 1 \ \forall s$);
   b) *Proposed*: the proposed training strategy, which modifies the vanilla strategy by correcting the value of the loss function taking in consideration the (estimated) transition matrices $\hat{\mathbf{T}}^{(s)}$ of the different sources as in (11).
   c) *Forward*: the loss correction approach [17], which is similar to *proposed* but uses a single transition matrix estimated on the entire training set, thus not exploiting the multi-source nature of the training labels.

2) Loss function (see (3))
   a) the standard *Categorical Cross Entropy* (CCE);
   b) the *Generalized Cross Entropy* (GCE) [11] with the default value for the hyperparameter $q = 0.7$.
   c) the *Symmetric Learning* (SL) [10] with the default values for its hyperparameters.

The general structure of an experiment is as follows. For each sampling of the dataset, a baseline classifier is trained on the clean dataset $\mathcal{D}_{train}^{(0)}$ with the vanilla training strategy adopting one of the loss functions considered. Then, for each sampling, the related baseline classifier is used to estimate the transition matrices $\hat{\mathbf{T}}^{(s)}$ as described in Section III-C. Next, the combinations of training strategy and loss functions are used to train the model, adopting the estimated transition matrices $\hat{\mathbf{T}}^{(s)}$ for the *proposed* and *forward* training techniques.

The design of the experiments revolved around the performances evaluation in terms of error rate, quantity and quality of sources, and amount of available data. We primarily studied these results on the EuroSAT dataset, and then made some key experiments on the NWPU-RESISC45 to confirm the findings and the generality of the proposed strategy. For the EuroSAT dataset, we focused on the case of class-dependent errors and designed several transition matrix templates (Fig. 2) where the support of each row is fixed while the error rate can change. This choice allowed us to study the performance as a function of the error rate for specific patterns of the errors (i.e., class-dependent and uniform errors). On the NWPU-RESISC45 dataset, we adopted transition matrices coherent with the transitions defined in [15].

TABLE II
MEAN OVERALL CLASSIFICATION ACCURACIES (%) AND RELATED STANDARD DEVIATION (IN BRACKETS) ON THREE RUNS WITH A SINGLE WEAK SOURCE IN ADDITION TO THE CLEAN ONE, BEST RESULTS WITHIN A ROW ARE IN BOLD (EUROSAT DATASET)

| Weak source[a] | Error rate $\eta$ | Vanilla CCE | Proposed CCE | Vanilla GCE | Proposed GCE |
|---|---|---|---|---|---|
| ×0 | – | 87.83 (1.15) | – | **88.49 (1.02)** | – |
| ×3 | 0.0 | 95.44 (0.61) | – | **95.56 (0.25)** | – |
|  | 0.1 | 91.31 (0.78) | 93.52 (0.69) | 93.93 (0.31) | **94.19 (0.38)** |
|  | 0.2 | 87.85 (0.99) | 92.72 (0.18) | 91.83 (0.38) | **93.43 (0.54)** |
|  | 0.3 | 85.41 (3.01) | 92.32 (0.16) | 88.38 (0.51) | **92.82 (0.80)** |
|  | 0.4 | 79.94 (0.62) | 92.62 (0.33) | 82.91 (1.26) | **93.12 (0.24)** |
|  | 0.5 | 71.00 (3.72) | 92.68 (0.48) | 71.09 (3.65) | **92.88 (0.44)** |
| ×6 | 0.0 | 96.45 (0.44) | – | **96.81 (0.38)** | – |
|  | 0.1 | 92.80 (0.31) | 94.79 (0.52) | 95.04 (0.37) | **95.60 (0.44)** |
|  | 0.2 | 91.50 (0.68) | 93.66 (0.19) | 93.11 (0.12) | **94.90 (0.54)** |
|  | 0.3 | 87.94 (1.32) | 93.49 (0.38) | 90.96 (1.17) | **94.07 (0.06)** |
|  | 0.4 | 79.65 (3.02) | 93.84 (0.31) | 82.91 (1.16) | **94.04 (0.57)** |
|  | 0.5 | 69.02 (4.40) | 93.80 (0.18) | 65.73 (1.82) | **94.30 (0.53)** |
| ×9 | 0.0 | 97.14 (0.34) | – | **97.38 (0.29)** | – |
|  | 0.1 | 94.44 (0.52) | 95.66 (0.43) | 95.96 (0.38) | **96.30 (0.46)** |
|  | 0.2 | 92.67 (0.52) | 94.68 (0.35) | 94.19 (0.61) | **95.98 (0.32)** |
|  | 0.3 | 90.75 (0.39) | 94.65 (0.24) | 91.46 (0.24) | **94.85 (0.39)** |
|  | 0.4 | 81.55 (1.18) | **94.90 (0.51)** | 81.90 (2.91) | 94.76 (0.41) |
|  | 0.5 | 63.60 (7.10) | 94.56 (0.26) | 61.60 (3.50) | **94.91 (0.22)** |

[a] "Weak source" refers to number of times the weak labeled data is larger than the clean one. "×0" means no weak labeled data are used (i.e., it refers to the baseline classifiers).

## V. EXPERIMENTAL RESULTS: EUROSAT DATASET

In our experiments, different numbers and kinds of weak sources were considered. In order to analyze the properties of the proposed strategy in detail, we first focused on the EuroSAT dataset and the case of a single weak source in addition to the clean one. In this way it was possible to study the effects of a large number of variables on the performance of the proposed technique. Then, we analyzed a significant case of multiple weak sources to confirm the findings of the case of a single weak source. Finally, we developed some specific experiments for analyzing the robustness of the proposed approach to different operational conditions.

In all the experiments on EuroSAT, the clean dataset was defined with $m^{(0)} = 0.05 \times m_{train}$ samples (~100 instances/class before data augmentation), where $m_{train} = 0.8 \cdot m$ is the number of instances in $\mathcal{D}_{train}$ (see Algorithm 1). Then, we primarily considered error rates $\eta = 0.1, 0.2, 0.3, 0.4, 0.5$ and weak sources with number of labeled instances $m^{(s)} = \{3, 6, 9\} \times m^{(0)}$.

### A. Case I: One weak source

Let us first consider the case of a single weak source (i.e., $S = 1$) in addition to the clean one on the EuroSAT dataset. The weak source was studied with $m^{(1)} = \{3, 6, 9\} \times m^{(0)}$ samples and the transition matrix template shown in Fig. 2(a). The matrix was designed w.r.t. a realistic weak source suffering from interclass similarities and label obsolescence (i.e., land-cover changes). Note that the two water body classes are left unchanged since their recognition is generally less prone to errors and they are less likely to change over time.

Table II shows the average of the best OAs on the test set achieved by training the ResNet-50 with four combinations of training strategies and loss functions (i.e., *vanilla* and *proposed* with CCE and GCE). The trained baseline classifiers overfitted the clean dataset, achieving an OA on the test set that hardly surpassed 90%. As anticipated, such classifiers were exploited to estimate the transition matrices used by the proposed training approach. The results show that the proposed training strategy outperforms the vanilla counterpart in all the cases of error rate. Figg. 3, 4 and 5 show the results of Table II from different perspectives. Let us analyze them.

*1) Error rate:* Fig. 3(a), 3(b) and 3(c) show that the proposed training strategy is always able to increase the OA w.r.t. the baseline classifiers, whereas the vanilla counterpart is strongly affected by high error rates. Note that for $\eta < 0.4$ the vanilla training strategy is still able to obtain increased accuracies. This can be understood in terms of model that exploits the majority voting assumption. Indeed, one can note that when the constraint (20) is not satisfied (i.e., $\eta \geq 0.4$) the trained models always achieve results worse than the baseline classifiers.

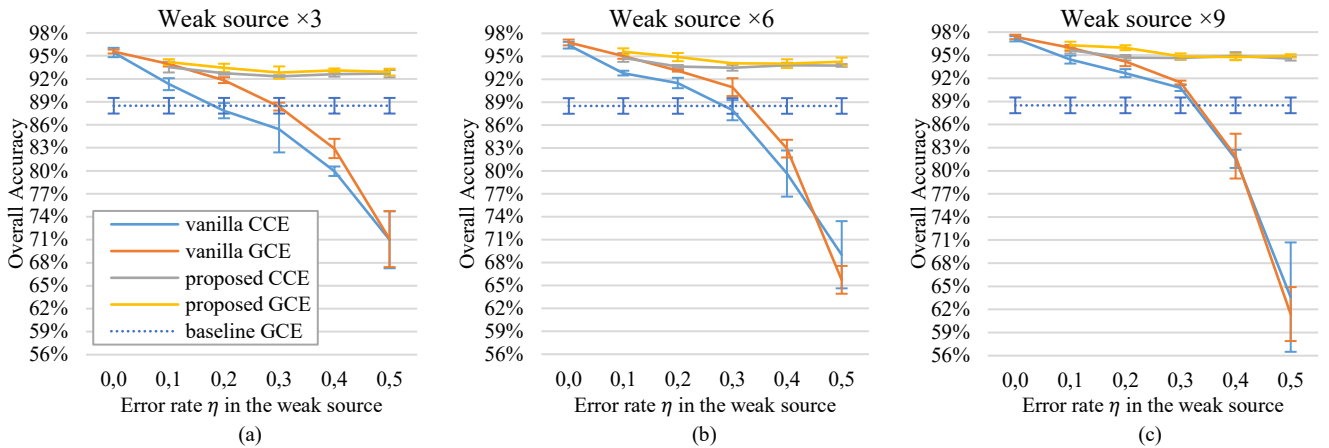

Fig. 3. Overall Accuracy ($S = 1$ weak source) vs. the error rate in the weak source. (a), (b) and (c) show how the Overall Accuracy changes as the error rate increases, in the three cases of $m^{(1)} = \{3, 6, 9\} \times m^{(0)}$, respectively. (EuroSAT dataset)

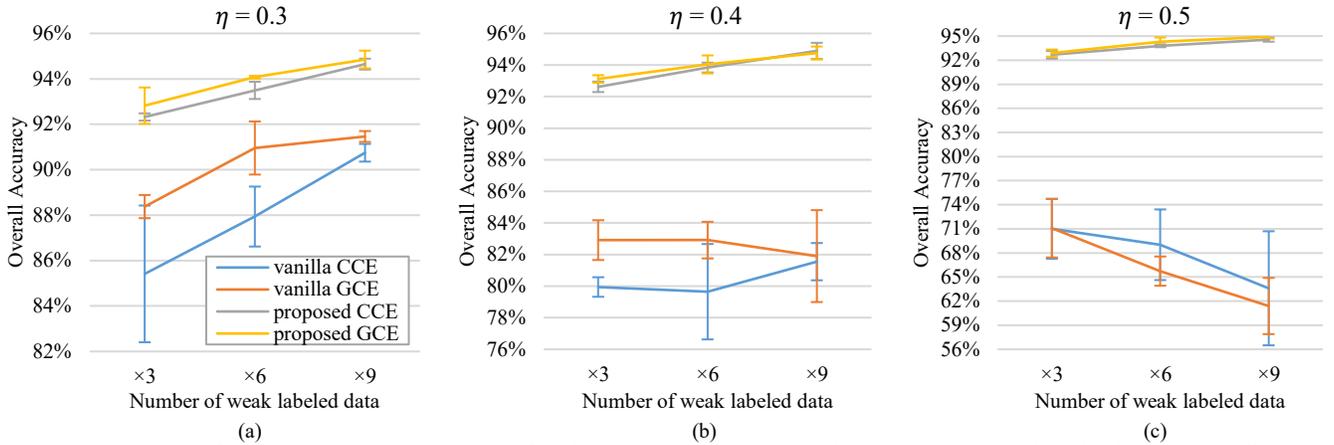

Fig. 4. Overall Accuracy ($S = 1$ weak source) vs. the quantity of weak labeled data w.r.t. the clean dataset. (a), (b) and (c) show the change in Overall Accuracy as $m^{(1)}$ increases in the three cases of error rate $\eta = 0.3, 0.4, 0.5,$ respectively. (EuroSAT dataset)

*2) Number of instances:* Fig. 4(a), 4(b) and 4(c) show the behavior of the OA on the test set vs. the number of weak labeled data for different amounts of error. One can note that the proposed approach is always capable of leveraging on the added weak labeled data regardless of the error rate for increasing the OA, whereas the vanilla counterpart can do the same only when the constraint (20) is satisfied (i.e., $\eta < 0.4$).

*3) Loss functions:* The results show that the GCE outperforms the CCE both in the vanilla and the proposed training strategy. In particular, when combined with the proposed training strategy, the GCE merges its robustness with the effectiveness of the proposed technique.

*4) Training dynamics:* Fig. 5 shows the behavior of the OA on the test set vs. the number of epochs in the training with $m^{(1)} = 9 \times m^{(0)}$. One can see that higher error rates in the weak sources cause a higher degradation of the accuracy over time. This degradation is sharply reduced by the proposed strategy and slightly reduced by the robust GCE. In particular, the proposed approach combined with GCE is the most robust against errors in the weak labeled data and reaches higher accuracies, thus successfully combining the benefits of the two techniques.

*5) Class-dependent vs. uniform errors:* In the theoretical analysis of Section III-E we discussed the increased difficulty of training with the proposed strategy under uniform errors. Hence, we decided to analyze this phenomenon in greater detail. We considered a weak source six times larger than the clean one in the two cases of transition matrix characterized by class-dependent (Fig. 2(a)) and uniform errors (Fig. 2(b)). Figg. 6(a) and 6(b) compares the obtained results, showing that class-dependent errors are indeed easier to handle. Moreover, one can observe a similarity between the results and the entropy of the related transition matrices (Fig. 6(c)), especially in the case of CCE (Fig. 6(a)). Indeed, in both the cases of uniform and class-dependent errors the proposed strategy shows a mirrored behavior w.r.t. the entropy, with better performance for low entropies and poorer performances for high entropies. The situation changes in the case of GCE (Fig. 6(b)), where in the case of class-dependent errors with $\eta \geq 0.7$ the performances drop and become more unstable. This could be related to the violation of the constraint (20). Indeed, combining a robust loss function as GCE with the proposed technique allows the optimizer to give more faith to confident predictions even when this is not supported by the transition matrix. In a situation where the majority of the labels are assigned to a wrong class, this behavior may lead to give too much trust to the given labels, undermining the learning process. This suggests that robust loss functions are useful for enhancing the performances only when dealing with low error rates, whereas they become unreliable for high error rates (i.e., when violating constraint (20)).

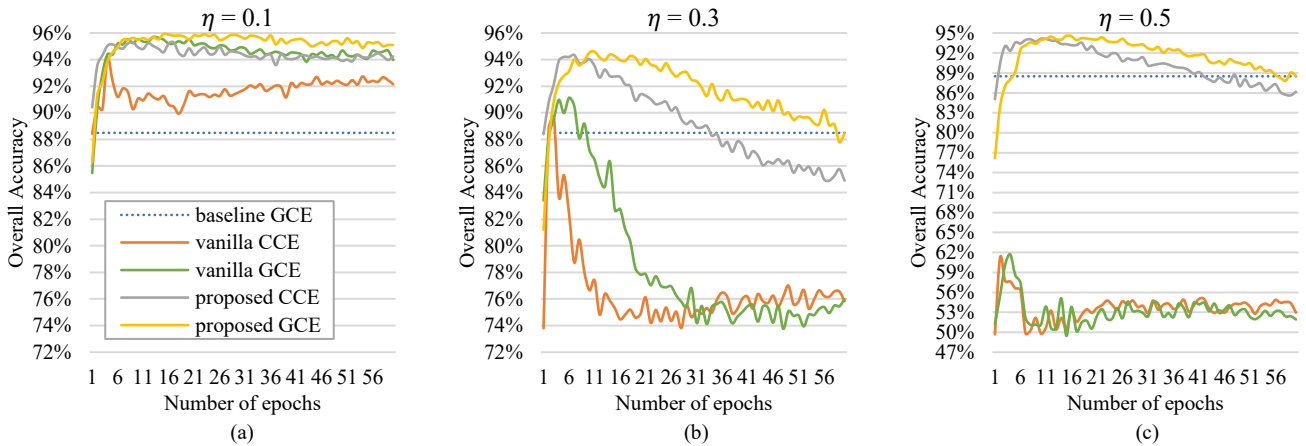

Fig. 5. Behavior of the Overall Accuracy on the test set vs. the number of epochs during the training in three cases of error rates (i.e., $\eta = 0.1, 0.3, 0.5$, respectively) in the weak source (case of one weak source nine times larger than the clean labeled dataset). (EuroSAT dataset)

TABLE III
MEAN OVERALL CLASSIFICATION ACCURACIES (%) AND RELATED STANDARD DEVIATION (IN BRACKETS) ON THREE RUNS WITH THREE WEAK SOURCES IN ADDITION TO THE CLEAN ONE, BEST RESULTS WITHIN A ROW ARE IN BOLD (EUROSAT DATASET)

| Weak sources[a] | Error rate $\eta$ | Vanilla CCE | Proposed CCE | Vanilla GCE | Proposed GCE |
|---|---|---|---|---|---|
| ×0 | – | 87.96 (0.87) | – | 89.44 (0.40) | – |
| ×9 | 0.1 | 94.08 (0.31) | 95.79 (0.13) | 96.26 (0.17) | **96.52 (0.82)** |
|  | 0.2 | 93.19 (0.60) | 95.33 (0.16) | 95.41 (0.23) | **95.73 (0.28)** |
|  | 0.3 | 91.64 (1.26) | 94.93 (0.37) | 93.96 (0.71) | **95.31 (0.29)** |
|  | 0.4 | 88.60 (1.26) | 94.12 (0.22) | 92.30 (0.53) | **94.78 (0.17)** |
|  | 0.5 | 82.16 (0.25) | 94.00 (0.26) | 85.59 (1.81) | **94.08 (0.18)** |

[a] "Weak sources" refers to number of times the total weak labeled data is larger than the clean ones. "×0" means no weak labeled data are used (i.e., it refers to the baseline classifiers).

### B. Case II: Three weak sources

Table III refers to the case of multiple weak sources (i.e., $S = 3$) in addition to the clean one. In this case, all the weak sources have been defined with the same amount of labeled instances, given by $m^{(s)} = 3 \times m^{(0)}$, $s = 1,2,3$, for a total amount of weak labeled data nine times larger than the clean dataset. They were characterized by different types of errors modelled by three different transition matrices:

1) The first source was characterized by uniform errors (Fig. 2(b)). Even if uncommon, we decided to consider it in our analysis.
2) One of the weak sources was characterized by class-dependent errors designed to mimic an obsolete source affected by land-cover changes (Fig. 2(c)).
3) Another source was also characterized by class-dependent errors, while instead designed to mimic an inaccurate source affected by interclass similarities (Fig. 2(d)).

The balanced error rate $\eta$ was set equal for all the weak sources and thus used to define all the three transition matrices.

From Table III one can see that, as in the previous case, the proposed approach obtains the best OA in all the cases. In particular, we observed that the OA of the proposed training approach is still slightly affected by the increasing error rate, whereas the OA of the vanilla training strategy shows a higher degradation for higher error rates. Hence, this behavior confirms the ability of the proposed approach to combine and exploit the weak sources.

By analyzing the results in greater detail, some differences can be observed comparing Table II (for the case of a weak source nine times larger than the clean one) with Table III, especially for the vanilla approaches. For simplicity, we plotted them in Fig. 7, which shows the OA of the vanilla strategies vs the error rate in the weak sources. One of the differences is that, in the case of three weak sources, the vanilla approaches reached OAs higher than before. Especially for CCE, this interesting result can be understood from the fact that now we are using multiple weak sources with different conditions for satisfying the constraint (20). From the transition matrices in Fig. 2(b), 2(c) and 2(d) one can see that (20) is met for values of $\eta$ smaller than 0.9, 0.3, and 0.5, respectively. Hence, since the error rates considered were always below or equal to 0.5, only the second (Fig. 2(c)) and the third (Fig. 2(d)) weak sources happen to violate the constraint at some point. Consequently, the model was always able to learn under the majority voting assumption from all the sources for $\eta < 0.3$. Instead, when $\eta = \{0.3, 0.4\}$ the model was able to learn under the majority voting assumption from all the sources except the second weak source, which comprises 3/10 of the multisource dataset $\mathcal{D}_{train}^{MS}$. When $\eta = 0.5$, the model was able to learn under the majority voting assumption only from the clean source and the first weak source (characterized by uniform errors), while being simultaneously misled by the remaining weak sources (6/10 of the multisource dataset $\mathcal{D}_{train}^{MS}$) that violated the constraint (20). For comparison, in the results of Table II with a single weak source nine times larger than the clean one, 9/10 of the multisource dataset $\mathcal{D}_{train}^{MS}$ is violating the constraint when $\eta \geq 0.4$. Therefore, these differences in terms of violations of the constraint (20) justify the generally increased performances (w.r.t. Case I) of the vanilla training strategy combined with CCE for $\eta \geq 0.4$.

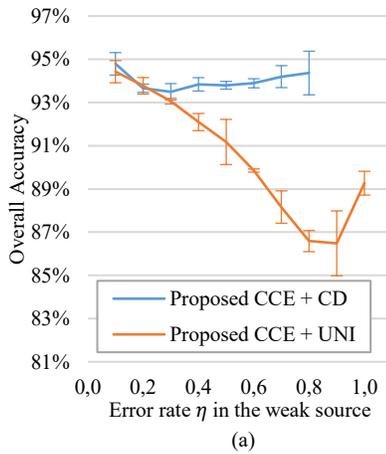 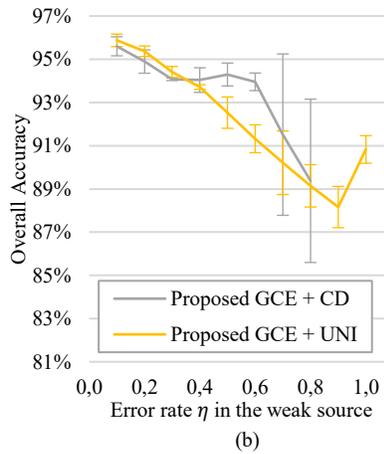 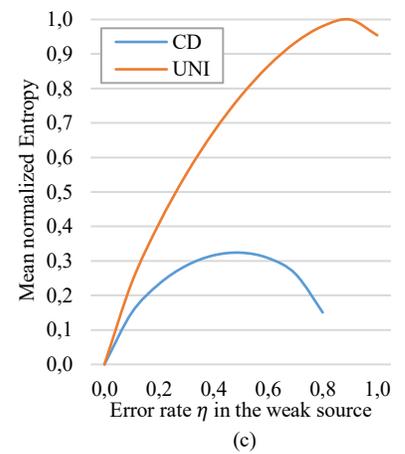

(a) (b) (c)

Fig. 6. (a) Overall Accuracy of the proposed training strategy (combined with CCE) with a single weak source affected by class-dependent (CD) errors (Fig. 2(a)) and uniform (UNI) errors (Fig. 2(b)) vs. the error rate. (b) Overall Accuracy of the proposed training strategy (combined with GCE) with a single weak source affected by class-dependent (CD) errors (Fig. 2(a)) and uniform (UNI) errors (Fig. 2(b)) vs. the error rate. (c) Entropy of the transition matrices in the two cases of class-dependent (CD) errors (Fig. 2(a)) and uniform (UNI) errors (Fig. 2(b)) vs. the error rate. Clearly, the performances are related to the informativeness of the transition matrices, and thus uniform errors are more difficult to recover since they are less informative (higher entropy).

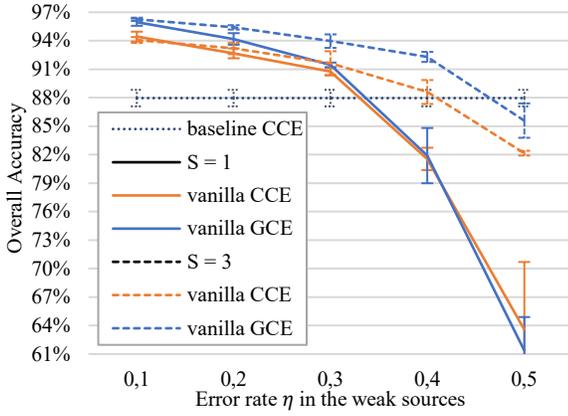

Fig. 7. Comparison of the OA vs. the error rate in the weak sources between Case I (i.e., $S=1$, one weak source nine times larger than the clean source) and Case II (i.e., $S=3$). (EuroSAT dataset)

Regarding the vanilla strategy combined with GCE, we observed a clear increase in term of robustness. This can be understood from two facts: 1) GCE is reported to be more effective against uniform errors than class-dependent errors [11], hence the presence of a weak source characterized by uniform errors partially justifies the increased OAs; 2) the presence of multiple different weak sources may make more difficult for the model to overfit the weak labels, making GCE capable to better sift out the wrong labels.

We observed another slight difference w.r.t. the case of one weak source regarding the proposed training strategy. Despite reaching similar OAs at low noise rates, the model trained in the case of three weak sources reached lower OAs at high noise rates. This is due to the presence of a weak source affected by uniform errors. Indeed, the higher entropy poses a more difficult optimization scenario, as confirmed by the experimental results in Fig. 6. Note that techniques that are unaware of the transition matrices (e.g., the vanilla training strategy combined with GCE) may find uniform errors easier to handle, whereas techniques that exploit the knowledge of the transition matrices prefer class-dependent errors, since they are more informative.

### C. Case III: Removing the clean source from Case I

One can note that when the weak source is nine times larger than the clean source, the clean labeled data comprises only 1/10 of the data used for training. Thus, one could argue that the clean dataset may be neglectable for the sake of the training of a performant classifier. For example, one could decide to use the clean labeled data to train a baseline classifier, which is used for the estimation of the transition matrix of the weak source, and then train the final classifier using only the weak labeled data and a loss correction approach like in [17]. However, the presence of a clean dataset during the training is fundamental, especially when combining with robust loss function like GCE. Fig. 8 shows the behavior of the OA vs. the number of epochs when discarding the clean dataset during the training (note that for better understanding the behavior of the method the transition matrices are still estimated using the baseline classifier trained with the clean source). In the case of CCE, the best OA achieved with the proposed training strategy when removing the clean source is close to the OA achieved when instead using it. However, the proposed training strategy shows to be able to exploit the clean dataset to alleviate the degradation problem w.r.t. the number of epochs and consequently to consistently reach better OAs. Instead, in the case of GCE, the combination of GCE with the proposed training strategy showed poor results when no clean labeled data were used. Even the vanilla strategy obtained better OAs without the clean dataset. On the contrary, when using also the clean source, the proposed training strategy combined with GCE outperformed all the other strategies reaching the best OAs. Hence, the few clean labeled data showed to be enough to allow the proposed training method to both exploit the weak sources and leverage on the most reliable data offered by the multisource dataset.

### D. Effects of the transition matrix estimation accuracy on the proposed technique

A requirement for the proposed training strategy is the estimation of the transition matrices characterizing the different weak sources. Here, this is done by training the DL model on the clean dataset $\mathcal{D}_{train}^{(0)}$, obtaining a baseline classifier useful for the estimation of the transition probabilities. As stated before and proved by the experimental results, approximate estimations of the transition matrices are enough to catch the underlying relationships between clean and weak labels of a specific source. Here, we better study the effects of the accuracy of the estimates on the final results.

We selected a single sampling of the EuroSAT dataset (i.e., $\mathcal{D}_{train}^{MS}$ and $\mathcal{D}_{test}$ are always the same) and combined the proposed training strategy with transition matrices estimated by three baseline classifiers with different OAs on the test set selected from the trained models saved during the training on

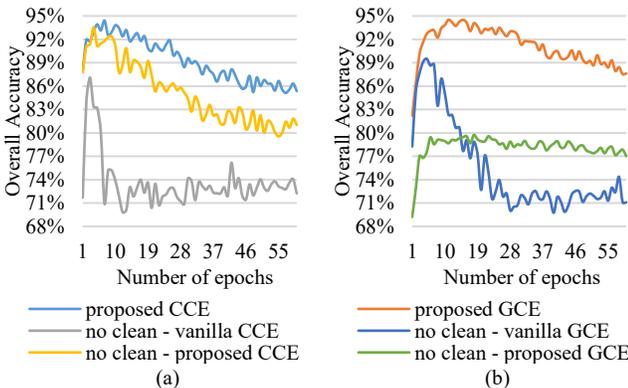

Fig. 8. Behavior of the Overall Accuracy on the test set vs. the number of epochs during the training with or without the clean source ($S = 1$, $m^{(1)} = 9 \times m^{(0)}$ and $\eta = 0.3$ with transition matrix as in Fig. 2(a)). (a) Results obtained with CCE. (b) Results obtained with GCE. (EuroSAT dataset)

TABLE IV
OVERALL ACCURACIES OF THE VANILLA APPROACH COMPARED TO THE PROPOSED APPROACH IN THE CASE OF DIFFERENT OAS OF THE BASELINE CLASSIFIERS USED FOR THE ESTIMATION OF THE TRANSITION MATRIX (EUROSAT DATASET)

| Error rate $\eta$ | OA (%) in the estimation of the transition matrix | Overall Accuracy (%) | |
|---|---|---|---|
| | | Proposed GCE | Vanilla GCE |
| 0.2 | 65.28 | 94.41 | |
| | 78.65 | 94.72 | 93.24 |
| | 89.91 | 95.33 | |
| 0.4 | 65.28 | 92.41 | |
| | 78.65 | 94.69 | 81.22 |
| | 89.91 | 94.44 | |

Case of $S = 1$ weak source six times larger than the clean one and with transition matrix as in Fig. 2(a).

TABLE V
COMPARISON OF THE PROPOSED STRATEGY WITH THE SoA METHODS. MEAN OVERALL CLASSIFICATION ACCURACIES (%) AND RELATED STANDARD DEVIATION (IN BRACKETS) WITH ONE WEAK SOURCE NINE TIMES LARGER THAN THE CLEAN ONE, BEST RESULTS WITHIN A ROW ARE IN BOLD (EUROSAT DATASET)

| Err. rate | CCE | SL | GCE | Forward | Proposed CCE |
|---|---|---|---|---|---|
| 0.1 | 94.44 (0.52) | 95.93 (0.48) | **95.96 (0.38)** | 95.56 (0.18) | 95.66 (0.43) |
| 0.2 | 92.67 (0.52) | 94.41 (0.26) | 94.19 (0.61) | 94.55 (0.52) | **94.68 (0.35)** |
| 0.3 | 90.75 (0.39) | 92.36 (1.23) | 91.46 (0.24) | 93.83 (0.26) | **94.65 (0.24)** |
| 0.4 | 81.55 (1.18) | 86.65 (0.89) | 81.90 (2.91) | 90.75 (3.01) | **94.90 (0.51)** |
| 0.5 | 63.60 (7.10) | 60.53 (1.03) | 61.60 (3.50) | 89.33 (2.63) | **94.56 (0.26)** |

the clean dataset. Specifically, we selected the saved models of the best epoch (OA of ~90%) and of two previous epochs with an OA of ~65% and ~79%, respectively. Then, we considered the case of a single weak source with transition matrix as in Fig. 2(a) labelling $m^{(1)} = 6 \times m^{(0)}$ images, in the two case of error rate $\eta$ values equal to 0.2 and 0.4. Table IV reports the resulting best OAs on the test set when training with both the vanilla strategy and the proposed strategy combined with transition matrices of different accuracy. GCE is used in all the cases. Note that the results of the vanilla strategy and of the proposed strategy with transition matrix estimated by a baseline classifier with an OA of ~90% are in line with the results of Table II. We observed that even a transition matrix characterized by an OA equal to ~65% is sufficient for reaching good performances (with a decrease of OA of 1–2% w.r.t. the best ones). For example, when $\eta = 0.4$, the proposed approach combined with such estimated transition matrix increased the OA on the test set of more than 10% w.r.t. the vanilla counterpart and of ~2% w.r.t. the best baseline classifier. Instead, when $\eta = 0.2$, the proposed approach showed an increase of the OA on the test set of ~1% w.r.t. the vanilla counterpart and of ~4% w.r.t. the best baseline classifier. These results show that accurate estimations of the transition matrices are not essential in our proposed strategy to exploit the available weak sources. The important requirement is to estimate a transition matrix that captures, even approximately, the underlying relationships between true and weak labels.

*E. Comparisons with the State of the Art methods*

Up to our knowledge, this is the first work studying the label noise problem under the multi-source labels setting. This makes comparisons with the SoA more challenging. For this reason, we decided to compare the proposed strategy (in the standard form, i.e. combined with CCE) with noise robust loss functions (i.e., GCE and SL) and the *forward* training strategy by treating the multisource labels as a single source. In particular, for the *forward* training strategy a single transition matrix is estimated for the entire training set with the trained baseline classifier. Table V shows the results with a weak source nine times larger than the clean one. Note that the results for CCE, GCE and the proposed strategy are the same of Table II, where the vanilla training strategy is implied when not specified. One can see that SL obtains worse results than GCE and that both are better than

CCE only for low error rates (i.e., when $\eta < 0.4$ and constraint (20) is satisfied). The *forward* strategy obtained similar results to the ones of other SoA methods at low error rates whereas showed to be more robust at higher error rates. This is related to the fact that the *forward* correction step helps to circumvent the violation of the constraint (20), just as in the proposed technique. However, explicitly modelling the presence of some clean labeled data as done in the proposed strategy allowed to achieve much better and more stable (i.e., lower standard deviation of the OA) results even for high error rates. These results suggest that also a scenario characterized by a single weak source could benefit from a multi-source formulation if some labeled data can be cleaned to form a small clean labeled dataset.

VI. EXPERIMENTAL RESULTS: NWPU-RESISC45 DATASET

As previously stated, on the NWPU-RESISC45 dataset we adopted transition matrices coherent with the transitions defined in [15]. As with the EuroSAT dataset, the matrix was defined as function of the error rate, hence the original labels are preserved with a probability of $1-\eta$ and the remaining probability values are proportionally modified. In all the experiments, the clean dataset was defined with $m^{(0)} = 0.1 \times m_{train}$ samples (56 instances/class), where $m_{train} = 0.8 \cdot m$ is the number of instances in $\mathcal{D}_{train}$ (see Algorithm 1). Then, we primarily considered error rates $\eta = 0.1, 0.3, 0.5, 0.7, 0.9$ and weak sources with a number of labeled instances given by $m^{(s)} = 9 \times m^{(0)}$.

Table VI shows the results of the proposed strategy combined with CCE compared with those of other SoA techniques. The baselines used to estimate the transition matrices have an OA of 79.37(0.76)% and 76.78(0.70)% in the case of ResNet-50 and VGG16, respectively. The results confirm the robustness of the proposed technique to any error rate. In particular, it always obtains the best results for the highest error rates and competing results for small error rates. The ResNet-50 showed better results w.r.t. the VGG16 model. Apart from that, the main difference between the two models is the behavior of the *forward* strategy. While in the case of ResNet-50 it shows highly degrading performances as the error rate increases, in the case of VGG16 it mitigates this behavior, showing competing results. This could be due to the use of regularization techniques such as L2-norm weight decay, dropout and batch normalization before the softmax function, which may have reinforced the generalization capabilities of the method.

However, note that the proposed strategy always shows better or similar accuracies, which are also more stable. As with the EuroSAT dataset, the increased stability is due to the multisource formulation, where the clean source helps by guiding the optimization process (which becomes more relevant at high error rates).

Figg. 9(a), 9(b) and 9(c) show the plots of the OAs presented in Tables V and VI. One can see that the results on the NWPU-RESISC45 dataset share a similar behavior with the results on the EuroSAT dataset. For example, one can note that the transition matrix used for the weak source in the NWPU-RESISC45 dataset violates the constraint (20) when $\eta \geq 5/9 \simeq 0.56$ (recall that $\eta \geq 0.4$ for the EuroSAT dataset with the transition matrix in Fig. 2(a)). From the slope of the curves it is possible to observe that in such case all the *vanilla* methods are strongly affected by the error rate and obtain worse results than the baseline, whereas the *forward* and the *proposed* strategies are only slightly affected by the error rate.

## VII. CONCLUSION

In remote sensing image scene classification, training DNNs having both high accuracy and good generalization capabilities is critical due to the scarcity of reliable training data. To solve this problem, in this paper we proposed the use of multiple sources of weak labeled data (e.g., obsolete and inaccurate digital maps) in addition to the few reliable training data. Starting from that, we presented a novel multisource training strategy that weighs and exploits each source by means of approximate estimations of the transition matrices characterizing the errors produced by them. The main idea exploited is to weight and use a training label to optimize multiple classes at time based on the approximate knowledge of the source originating it. This allows the model to properly optimize classes also with mislabeled samples, while weighting them in order to let the most reliable labels guide the optimization process. The proposed method, whose properties have been studied from the theoretical perspective, is general, simple to implement and only requires the availability of a small reliable dataset, which is usually available in supervised classification problems.

TABLE VI
MEAN OVERALL CLASSIFICATION ACCURACIES (%) AND RELATED STANDARD DEVIATION (IN BRACKETS) ON NWPU-RESISC45 WITH ONE WEAK SOURCE NINE TIMES LARGER THAN THE CLEAN ONE, BEST RESULTS WITHIN A ROW ARE IN BOLD

| Err. rate | CCE | SL | GCE | Forward | Proposed CCE |
|---|---|---|---|---|---|
| ResNet-50 | | | | | |
| 0.1 | 84.07 (0.29) | 85.81 (0.34) | **86.95 (0.56)** | 86.05 (0.51) | 86.47 (0.70) |
| 0.3 | 78.12 (0.28) | 84.53 (0.37) | 84.61 (0.82) | 82.25 (1.93) | **85.47 (0.33)** |
| 0.5 | 67.51 (0.75) | 79.78 (0.78) | 77.35 (1.05) | 71.51 (1.84) | **85.30 (0.32)** |
| 0.7 | 43.16 (1.25) | 56.25 (0.53) | 47.75 (1.89) | 58.99 (8.21) | **84.42 (0.37)** |
| 0.9 | 16.37 (0.37) | 16.41 (0.93) | 14.76 (0.28) | 58.99 (4.93) | **85.25 (0.50)** |
| VGG16 | | | | | |
| 0.1 | 83.88 (0.81) | 82.61 (0.41) | 83.57 (0.31) | **84.97 (0.80)** | 84.32 (0.74) |
| 0.3 | 80.57 (0.77) | 81.88 (0.40) | 82.04 (0.37) | 83.16 (0.43) | **83.65 (0.43)** |
| 0.5 | 72.88 (1.20) | 78.13 (0.89) | 78.42 (1.10) | **82.42 (0.80)** | 82.37 (0.11) |
| 0.7 | 50.75 (1.62) | 57.42 (0.35) | 58.55 (0.62) | 79.05 (1.92) | **82.21 (0.75)** |
| 0.9 | 17.14 (0.61) | 12.14 (0.38) | 12.54 (1.31) | 80.42 (1.11) | **82.61 (0.40)** |

The experimental results showed that the proposed strategy is able both to exploit the weak sources and to weigh them by leveraging on the most reliable labels, reaching always the best OAs on the test set with a large margin in the case of high error rates and competing results at lower rates. This shows that the use of robust training strategies like the proposed one or the *forward* approach are generally more effective against high-level label noise than robust loss functions. Other experiments have proven that the main factor affecting the performances of the training strategy is the entropy of the different weak sources, thus making the uniform errors the worst case scenario for the proposed approach. Furthermore, the proposed strategy showed increasing OAs as the amount of weak labeled data increase and it also proved its generality as being successful with both pre-trained networks and models trained from scratch.

During the training, the OAs of the models showed a degradation behavior vs. the number of epochs after reaching its maximum. Hence, as future research we plan to analyze in-depth this behavior and to design strategies being able to mitigate it. Also, real-world weak labeled data may be dependent on both the true class and the underlying signal. Therefore, future experiments will also aim at the evaluation of the proposed strategy with real weak labeled data. Finally, the proposed strategy will be studied considering additional loss functions and different DL methodologies.

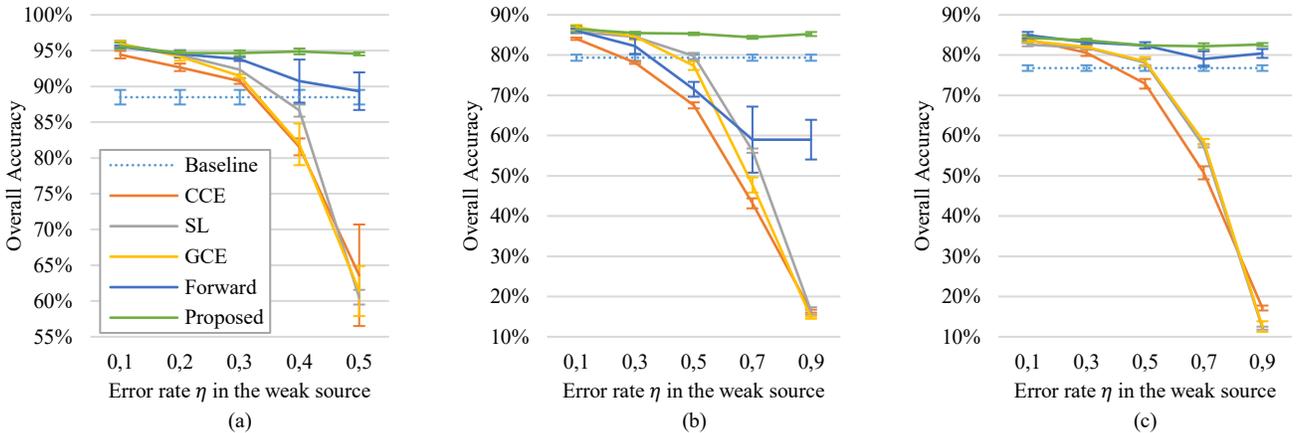

Fig. 9. Overall Accuracy vs. the error rate in the weak source ($m^{(1)} = 9 \times m^{(0)}$) in the different cases of benchmark dataset and model. (a) ResNet-50 trained on the EuroSAT dataset. (b) Pre-trained ResNet-50 + MLP trained on the NWPU-RESISC45 dataset. (c) Pre-trained VGG16 + MLP trained on the NWPU-RESISC45 dataset.

REFERENCES

[1] G. Cheng, X. Xie, J. Han, L. Guo and G.-S. Xia, "Remote sensing image scene classification meets deep learning: Challenges, methods, benchmarks, and opportunities," *IEEE J. Sel. Topics Appl. Earth Observ. Remote Sens.*, vol. 13, pp. 3735-3756, June 2020.

[2] L. Bruzzone, "Multisource labeled data: An opportunity for training deep learning networks," in *Proc. IEEE Int. Geosci. Remote Sens. Symp. (IGARSS)*, Yokohama, Japan, 2019, pp. 4799-4802.

[3] R. Hänsch and O. Hellwich, "The truth about ground truth: Label noise in human-generated reference data," in *Proc. IEEE Int. Geosci. Remote Sens. Symp. (IGARSS)*, Yokohama, Japan, 2019, pp. 5594-5597.

[4] A. Ghosh, H. Kumar and P. S. Sastry, "Robust loss functions under label noise for deep neural networks," in *Proc. 31st AAAI Conf. Artif. Intell.*, San Francisco, California, USA, 2017, pp. 1919–1925.

[5] J. Jiang, J. Ma, Z. Wang, C. Chen and X. Liu, "Hyperspectral image classification in the presence of noisy labels," *IEEE Trans. Geosci. Remote Sens.*, vol. 57, no. 2, pp. 851-865, Feb. 2019.

[6] Y. Li, Y. Zhang and Z. Zhu, "Error-Tolerant Deep Learning for Remote Sensing Image Scene Classification," *IEEE Trans. Cybern.*, vol. 51, no. 4, pp. 1756-1768, Apr. 2021.

[7] B. Tu, W. Kuang, W. He, G. Zhang and Y. Peng, "Robust learning of mislabeled training samples for remote sensing image scene classification," *IEEE J. Sel. Topics Appl. Earth Observ. Remote Sens.*, vol. 13, pp. 5623-5639, Sep. 2020.

[8] J. Wang, F. Gao, J. Dong and S. Wang, "Synthetic Aperture Radar Images Changes Detection based on Random Label Propagation," in *Proc. 10th Int. Workshop Anal. Multitemporal Remote Sens. Images (MultiTemp)*, Shanghai, China, 2019.

[9] B. Tu, C. Zhou, D. He, S. Huang and A. Plaza, "Hyperspectral classification with noisy label detection via superpixel-to-pixel weighting distance," *IEEE Trans. Geosci. Remote Sens.*, vol. 58, no. 6, pp. 4116-4131, Jan. 2020.

[10] Y. Wang, X. Ma, Z. Chen, Y. Luo, J. Yi and J. Bailey, "Symmetric cross entropy for robust learning with noisy labels," in *IEEE/CVF Int. Conf. Comput. Vis. (ICCV)*, Seoul, Korea (South), 2019, pp. 322-330.

[11] Z. Zhang and M. Sabuncu, "Generalized cross entropy loss for training deep neural networks with noisy labels," in *Proc. Adv. Neural Inf. Process. Syst. 31 (NeurIPS)*, Montréal, Canada, 2018, pp. 8778-8788.

[12] Y. Xu, P. Cao, Y. Kong and Y. Wang, "L_DMI: A novel information-theoretic loss function for training deep nets robust to label noise," in *Proc. Adv. Neural Inf. Process. Syst. 32 (NeurIPS)*, Vancouver, Canada, 2019, pp. 6225-6236.

[13] X. Wang, Y. Hua, E. Kodirov and N. M. Robertson, "IMAE for noise-robust learning: Mean absolute error does not treat examples equally and gradient magnitude's variance matters," 2020, arXiv:1903.12141 [cs.LG]. [Online]. Available: https://arxiv.org/abs/1903.12141.

[14] B. B. Damodaran, R. Flamary, V. Seguy and N. Courty, "An Entropic Optimal Transport loss for learning deep neural networks under label noise in remote sensing images," *J. Comput. Vis. Image Understand.*, vol. 191, no. 102863, 2020.

[15] J. Kang, R. Fernandez-Beltran, P. Duan, X. Kang and A. J. Plaza, "Robust Normalized Softmax Loss for Deep Metric Learning-Based Characterization of Remote Sensing Images With Label Noise," *IEEE Trans. Geosci. Remote Sens.*, pp. 1-14, Dec. 2020.

[16] J. Kang, R. Fernandez-Beltran, X. Kang, J. Ni and A. Plaza, "Noise-Tolerant Deep Neighborhood Embedding for Remotely Sensed Images With Label Noise," *IEEE J. Sel. Topics Appl. Earth Observ. Remote Sens.*, pp. 2551-2562, Feb. 2021.

[17] G. Patrini, A. Rozza, A. Menon, R. Nock and L. Qu, "Making deep neural networks robust to label noise: A loss correction approach," in *Proc. IEEE Conf. Comput. Vis. Pattern Recognit. (CVPR)*, Honolulu, HI, USA, 2017, pp. 2233-2241.

[18] P. Li, X. He, M. Qiao, X. Cheng, Z. Li, H. Luo, D. Song, D. Li, S. Hu, R. Li, P. Han, F. Qiu, H. Guo, J. Shang and Z. Tian, "Robust deep neural networks for road extraction from remote sensing images," *IEEE Trans. Geosci. Remote Sens.*, early access. doi: 10.1109/TGRS.2020.3023112.

[19] P. Ghamisi, B. Rasti, N. Yokoya, Q. Wang, B. Hofle, L. Bruzzone, F. Bovolo, M. Chi, K. Anders, R. Gloaguen, P. M. Atkinson and J. A. Benediktsson, "Multisource and multitemporal data fusion in remote sensing: A comprehensive review of the state of the art," *IEEE Geosci. Remote Sens. Mag.*, vol. 7, no. 1, pp. 6-39, Mar. 2019.

[20] P. Helber, B. Bischke, A. Dengel and D. Borth, "Introducing Eurosat: A Novel Dataset and Deep Learning Benchmark for Land Use and Land Cover Classification," in *Proc. IEEE Int. Geosci. Remote Sens. Symp. (IGARSS)*, Valencia, Spain, 2018.

[21] P. Helber, B. Bischke, A. Dengel and D. Borth, "EuroSAT: A novel dataset and deep learning benchmark for land use and land cover classification," *IEEE J. Sel. Topics Appl. Earth Observ. Remote Sens.*, vol. 12, no. 7, pp. 2217-2226, July 2019.

[22] K. He, X. Zhang, S. Ren and J. Sun, "Deep residual learning for image recognition," in *Proc. IEEE Conf. Comput. Vis. Pattern Recognit. (CVPR)*, Las Vegas, NV, USA, 2016, pp. 770-778.

[23] K. He, X. Zhang, S. Ren and J. Sun, "Identity mappings in deep residual networks," in *Proc. 14th Eur. Conf. Comput. Vis. (ECCV)*, Amsterdam, The Netherlands, 2016, pp. 630-645.

[24] G. Cheng, J. Han and X. Lu, "Remote Sensing Image Scene Classification: Benchmark and State of the Art," *Proc. IEEE*, vol. 105, no. 10, pp. 1865-1883, 2017.

[25] O. Russakovsky, J. Deng, H. Su, J. Krause, S. Satheesh, S. Ma, Z. Huang, A. Karpathy, A. Khosla, M. Bernstein, A. C. Berg and L. Fei-Fei, "ImageNet Large Scale Visual Recognition Challenge," *Int. J. Comput. Vis.*, vol. 115, no. 3, pp. 211-252, 2015.

[26] K. Simonyan and A. Zisserman, "Very Deep Convolutional Networks for Large-Scale Image Recognition," in *Proc. 3rd Int. Conf. Learn. Repr. (ICLR)*, San Diego, CA, USA, 2015.

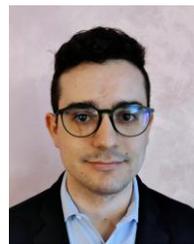

**Gianmarco Perantoni** received the "Laurea" (B.Sc.) degree in Information and Communication Engineering and the "Laurea Magistrale" (M.Sc.) degree in Information and Communication Engineering (summa cum laude) from the University of Trento, Italy, in 2018 and 2020, respectively.

He is currently a research fellow at RSLab in the Department of Information Engineering and Computer Science, University of Trento, Italy. His research interests concern the development of machine and deep learning techniques tailored to the analysis and processing of remotely sensed optical data.

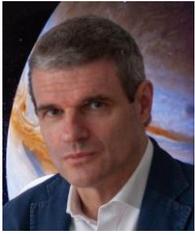

**Lorenzo Bruzzone** received the Laurea (M.S.) degree in electronic engineering (summa cum laude) and the Ph.D. degree in telecommunications from the University of Genoa, Italy, in 1993 and 1998, respectively. He is currently a Full Professor of telecommunications at the University of Trento, Italy, where he teaches remote sensing, radar, and digital communications.

Dr. Bruzzone is the founder and the director of the Remote Sensing Laboratory (https://rslab.disi.unitn.it/) in the Department of Information Engineering and Computer Science, University of Trento. His current research interests are in the areas of remote sensing, radar and SAR, signal processing, machine learning and pattern recognition. He promotes and supervises research on these topics within the frameworks of many national and international projects. He is the Principal Investigator of many research projects. Among the others, he is currently the Principal Investigator of the *Radar for icy Moon exploration (RIME)* instrument in the framework of the *JUpiter ICy moons Explorer (JUICE)* mission of the European Space Agency (ESA) and the Science Lead for the *High Resolution Land Cover* project in the framework of the Climate Change Initiative of ESA. He is the author (or coauthor) of 294 scientific publications in referred international journals (221 in IEEE journals), more than 340 papers in conference proceedings, and 22 book chapters. He is editor/co-editor of 18 books/conference proceedings and 1 scientific book. His papers are highly cited, as proven from the total number of citations (more than 38000) and the value of the h-index (91) (source: Google Scholar). He was invited as keynote speaker in more than 40 international conferences and workshops. Since 2009 he has been a member of the Administrative Committee of the IEEE Geoscience and Remote Sensing Society (GRSS), where since 2019 he is Vice-President for Professional Activities. Dr. Bruzzone ranked first place in the Student Prize Paper Competition of the 1998 IEEE International Geoscience and Remote Sensing Symposium (IGARSS), Seattle, July 1998. Since that he was recipient of many international and national honors and awards, including the recent *IEEE GRSS 2015 Outstanding Service Award*, the 2017 and 2018 *IEEE IGARSS Symposium Prize Paper Awards* and the *2019 WHISPER Outstanding Paper Award*. Dr. Bruzzone was a Guest Co-Editor of many Special Issues of international journals. He is the co-founder of the IEEE International Workshop on the Analysis of Multi-Temporal Remote-Sensing Images (MultiTemp) series and is currently a member of the Permanent Steering Committee of this series of workshops. Since 2003 he has been the Chair of the SPIE Conference on Image and Signal Processing for Remote Sensing. He has been the founder of the IEEE GEOSCIENCE AND REMOTE SENSING MAGAZINE for which he has been Editor-in-Chief between 2013-2017.Currently he is an Associate Editor for the IEEE TRANSACTIONS ON GEOSCIENCE AND REMOTE SENSING. He has been *Distinguished Speaker* of the IEEE Geoscience and Remote Sensing Society between 2012-2016. He is a Fellow of IEEE.